\documentclass{article} 

\usepackage{fullpage}
\usepackage{natbib}

\usepackage{hyperref}
\usepackage{url}

\usepackage{amsthm}
\usepackage{amsmath}
\usepackage{amsfonts}
\usepackage{graphicx}
\usepackage{subfig}
\usepackage{multirow}
\usepackage{wrapfig}
\usepackage{booktabs}
\usepackage{algorithm}
\usepackage[noend]{algpseudocode}
\newtheorem{theorem}{Theorem}

\title{
\bf{Plan Better Amid Conservatism: 
Offline Multi-Agent Reinforcement Learning with Actor Rectification
}
}

\date{}

\author{Ling Pan$^1$, Longbo Huang$^1$, Tengyu Ma$^2$, Huazhe Xu$^2$\\
$^1$Institute for Interdisciplinary Information Sciences, Tsinghua University \\
$^2$Stanford University \\
}

\begin{document}

\maketitle

\begin{abstract}
Conservatism has led to significant progress in offline reinforcement learning (RL) where an agent learns from pre-collected datasets. However, as many real-world scenarios involve interaction among multiple agents, it is important to resolve offline RL in the multi-agent setting. Given the recent success of transferring online RL algorithms to the multi-agent setting, one may expect that offline RL algorithms will also transfer to multi-agent settings directly. Surprisingly, we empirically observe that conservative offline RL algorithms do not work well in the multi-agent setting---the performance degrades significantly with an increasing number of agents. Towards mitigating the degradation, we identify a key issue that non-concavity of the value function makes the policy gradient improvements prone to local optima. Multiple agents exacerbate the problem severely, since the suboptimal policy by any agent can lead to uncoordinated global failure. Following this intuition, we propose a simple yet effective method, \underline{O}ffline \underline{M}ulti-Agent RL with \underline{A}ctor \underline{R}ectification (OMAR), which combines the first-order policy gradients and zeroth-order optimization methods to better optimize the conservative value functions over the actor parameters. Despite the simplicity, OMAR achieves state-of-the-art results in a variety of multi-agent control tasks. 
\end{abstract}

\section{Introduction}

Offline reinforcement learning (RL) has shown great potential in advancing the deployment of RL in real-world tasks where interaction with the environment is prohibitive, costly, or risky~\citep{thomas2015safe}.
Since an agent has to learn from a given pre-collected dataset in offline RL, it becomes challenging for online off-policy RL algorithms 
due to extrapolation error~\citep{fujimoto2019off,lee2021offline}.

There has been recent progress in tackling the problem based on conservatism. Behavior regularization~\citep{wu2019behavior,kumar2019stabilizing}, \textit{e.g.}, TD3 with Behavior Cloning (TD3+BC)~\citep{fujimoto2021minimalist}, compels the learning policy to stay close to the data manifold. Yet, its performance highly depends on the data quality.
Another line of research incorporates conservatism into the value function by critic regularization~\citep{nachum2019algaedice,kostrikov2021offline}, \textit{e.g.}, Conservative Q-Learning (CQL)~\citep{kumar2020conservative}, which usually learns a conservative estimate of the value function to directly address extrapolation error.

However, many practical scenarios involve multiple agents, \textit{e.g.}, multi-robot control~\citep{amato2018decision}, autonomous driving~\citep{pomerleau1989alvinn,sadigh2016planning}. Therefore, offline multi-agent reinforcement learning (MARL)~\citep{yang2021believe,jiang2021offline,mathieu2021starcraft} is crucial for solving real-world tasks. 
Recent results have shown that online RL algorithms can be applied to multi-agent scenarios through either decentralized training or a centralized value function without bells and whistles. For example, PPO~\citep{schulman2017proximal} leads to the effective methods Independent PPO~\citep{de2020independent} and Multi-Agent PPO~\citep{yu2021surprising} for the multi-agent setting.  Thus, we naturally expect that offline RL algorithms can also transfer easily when applied to multi-agent tasks.

Surprisingly, we find that the performance of the state-of-the-art conservatism-based CQL algorithm in offline RL degrades dramatically with an increasing number of agents, as shown in Figure \ref{fig:simple_navi}(b) in our experiments.
We demonstrate that actor optimization suffers from poor local optima, failing to leverage the global information in the conservative critics well. 
As a result, it leads to uncoordinated suboptimal learning behavior.
The issue is exacerbated severely with more agents and exponentially-sized joint action space~\citep{yang2021believe} in the offline setting, because the suboptimal policy of a single agent could lead to a global failure due to lack of coordination. 
For example, consider a basketball game where there are two competing teams each consisting of five players. When one of the players passes the ball among them, it is important for all the teammates to perform their duties well in their roles to win the game. As a result, if one of the agents in the team fails to learn a good policy, it can fail to cooperate with other agents for coordinated behaviors and lose the ball.

In this paper, we propose a simple yet effective method for offline multi-agent 
control, \underline{O}ffline \underline{M}ARL with \underline{A}ctor \underline{R}ectification (OMAR), to better leverage the conservative value functions.
Zeroth-order optimization methods, \textit{e.g.}, evolution strategies~\citep{such2017deep,conti2017improving,mania2018simple,salimans2017evolution}, recently emerged as another paradigm for solving decision making tasks that are robust to local optima, while this is usually not the case for first-order policy gradient methods~\citep{nachum2016improving,ge2017learning,safran2017spurious}.
Based on this inspiration, we introduce a new combination of the first-order policy gradient and the zeroth-order optimization methods in OMAR so that we can effectively combine the best of both worlds.
Towards this goal, in addition to the standard actor gradient update, we encourage the actor to mimic actions from the zeroth-order optimizer that maximize Q-values.
Specifically, the zeroth-order optimization part maintains an iteratively updated and refined sampling distribution to find better actions based on Q-values, where we propose an effective sampling mechanism.
We then rectify the policy towards these discovered better actions by adding a regularizer to the actor loss.

We conduct extensive experiments in standard continuous control multi-agent particle environments, the complex multi-agent locomotion task, and the challenging discrete control StarCraft II micromanagement benchmark to demonstrate its effectiveness. On all the benchmark tasks, OMAR significantly outperforms strong baselines, including the multi-agent version of current offline RL algorithms including CQL and TD3+BC, as well as a recent offline MARL algorithm MA-ICQ~\citep{yang2021believe}, and achieves the state-of-the-art performance. 

The main contribution of this work can be summarized as follows. 
We demonstrate the critical challenge of conservatism-based algorithms in the offline multi-agent setting empirically. We propose OMAR, a new algorithm to solve offline MARL tasks.
In addition, we theoretically prove that OMAR leads to safe policy improvement. Finally, we conduct extensive experiments to investigate the effectiveness of OMAR. Results show that OMAR significantly outperforms strong baseline methods and achieves state-of-the-art performance in standard continuous and discrete control tasks using offline datasets with different qualities. Videos are available at \url{https://sites.google.com/view/omar-videos}.
\section{Background}
Partially observable Markov games (POMG)~\citep{littman1994markov,hu1998multiagent} extend Markov decision processes to the multi-agent setting. A POMG with $N$ agents is defined by a set of global states $\mathcal{S}$, a set of actions $\mathcal{A}_1, \ldots, \mathcal{A}_N$, and a set of observations $\mathcal{O}_1, \ldots, \mathcal{O}_N$ for each agent. At each timestep, each agent $i$ receive an observation $o_i$ and chooses an action based on its policy $\pi_i$. The environment transits to the next state according to the state transition function $\mathcal{P}: \mathcal{S} \times \mathcal{A}_1 \times \ldots \times \mathcal{A}_N \times \mathcal{S} \to [0,1]$. Each agent receives a reward based on the reward function $r_i: \mathcal{S} \times \mathcal{A}_1 \ldots \times \mathcal{A}_N \to \mathbb{R}$ and a private observation $o_i: \mathcal{S} \to \mathcal{O}_i$. 
The goal is to find a set of optimal policies $\boldsymbol{\pi}=\{\pi_1, \ldots, \pi_N\}$, where each agent aims to maximize its own discounted return $\sum_{t=0}^{\infty} \gamma^t r_i^t$ with $\gamma$ denoting the discount factor. 
In the offline setting, agents learn from a fixed dataset $\mathcal{D}$ generated from the behavior policy $\boldsymbol{\pi}_{\beta}$ without interaction with the environments.

\subsection{Multi-Agent Actor Critic} \label{sec:maac}
\paragraph{Centralized critic.} 
\citet{lowe2017multi} propose Multi-Agent Deep Deterministic Policy Gradients (MADDPG) under the centralized training with decentralized execution (CTDE) paradigm by extending the DDPG algorithm~\citep{lillicrap2016continuous} to the multi-agent setting. In CTDE, agents are trained in a centralized way where they can access to extra global information during training while they need to learn decentralized policies in order to act based only on local observations during execution. In MADDPG, for an agent $i$, the centralized critic $Q_i$ is parameterized by $\theta_i$. It takes the global state action joint action as inputs, and aims to minimize the temporal difference error defined by 
$\mathcal{L}(\theta_{i})=\mathbb{E}_{\mathcal{D}}\left[({Q}_{i}\left(s, a_{1}, \ldots, a_{n}\right)-y_i\right)^{2}]$, where $y_i=r_i + \gamma \bar{Q}_i(s^{\prime}, a_{1}^{\prime}, \cdots, a_{n}^{\prime})|_{a_{j}^{\prime}=\bar{\pi}_{j}(o_{j}^{\prime})}$ and $\bar{Q}_i$ and $\bar{\pi}_i$ denote target networks. To reduce overestimation in MADDPG, MATD3~\citep{ackermann2019reducing} estimates the target value based on TD3~\citep{fujimoto2018addressing}, where $y_i=r_i + \gamma \min_{k=1,2} \bar{Q}_i^k(s^{\prime}, a_{1}^{\prime}, \cdots, a_{n}^{\prime})|_{a_{j}^{\prime}=\bar{\pi}_{j}(o_{j}^{\prime})}$.
Agents learn decentralized policies $\pi_i$ parameterized by $\phi_i$, which take only local observations as inputs. They are trained by multi-agent policy gradients according to $\nabla_{\phi_{i}} J(\pi_i) = \mathbb{E}_{\mathcal{D}}\left[\nabla_{\phi_i} \pi_i(a_{i} | o_i) \nabla_{a_i} Q_i\left(s, a_{1}, \ldots, a_{n}\right)|_{a_{i}=\pi_i(o_{i})}\right]$, where $a_i$ is predicted from its policy while $a_{-i}$ is sampled from the replay buffer.

\paragraph{Decentralized critic.}
Although centralized critics are widely-adopted in multi-agent methods, they lack scalability because the joint action space is exponentially large in the number of agents~\citep{iqbal2019actor}. On the other hand, independent learning approaches train decentralized critics that take only the local observation and action as inputs. It is shown in~\citet{de2020independent,lyu2021contrasting} that decentralized value functions can result in more robust performance and be beneficial in practice compared with centralized critic approaches. \citet{de2020independent} propose Independent Proximal Policy Optimization (IPPO) based on PPO~\citep{schulman2017proximal}, and show that it can match or even outperform CTDE approaches in the challenging discrete control benchmark tasks~\citep{samvelyan2019starcraft}. We can also obtain the Independent TD3 (ITD3) algorithm based on decentralized critics, which is trained to minimize the temporal difference error defined by $\mathcal{L}(\theta_{i})=\mathbb{E}_{\mathcal{D}_i}\left[\left(Q_{i}\left(o_i, a_i\right)-y_i\right)^{2}\right]$, where $y_i=r_i + \gamma \min_{k=1,2} \bar{Q}_i^k(o_i^{\prime}, \bar{\pi}_i(o_i^{\prime}))$.

\subsection{Conservative Q-Learning}
Conservative Q-Learning (CQL)~\citep{kumar2020conservative} adds a regularizer to the critic loss to address the extrapolation error and learns lower-bounded Q-values. It penalizes Q-values of state-action pairs sampled from a uniform distribution or a policy while encouraging Q-values for state-action pairs in the dataset to be large. Specifically, when built upon decentralized critic methods in MARL, the critic loss is defined as in Eq. (\ref{eq:cql}), where $\alpha$ is the regularization coefficient and $\hat{\pi}_{\beta_i}$ is the empirical behavior policy of agent $i$.
\begin{equation}
% \mathbb{E}_{\mathcal{D}_i} \left[ (Q_i(o_i,a_i)-y_i)^2 \right] 
\mathcal{L}(\theta_i) + \alpha \mathbb{E}_{\mathcal{D}_i}[\log \sum_{a_i} \exp (Q_i(o_i, a_i)) 
- \mathbb{E}_{a_i \sim \hat{\pi}_{\beta_i}}[Q_i(o_i, a_i)]]
\label{eq:cql}
\end{equation} 
\section{Proposed Method}
In this section, we first provide a motivating example where state-of-the-art offline RL methods, including CQL~\citep{kumar2020conservative} and TD3+BC~\citep{fujimoto2021minimalist}, can be inefficient in the face of the challenging multi-agent setting.
Then, we propose \underline{O}ffline \underline{M}ulti-Agent Reinforcement Learning with \underline{A}ctor \underline{R}ectification (OMAR), 
a simple yet effective method for the actors to better optimize the conservative value functions. 

\subsection{The Motivating Example} \label{sec:mot}
\begin{wrapfigure}{r}{0.35\textwidth} \vspace{-.2in}
\centering
\includegraphics[width=1.\linewidth]{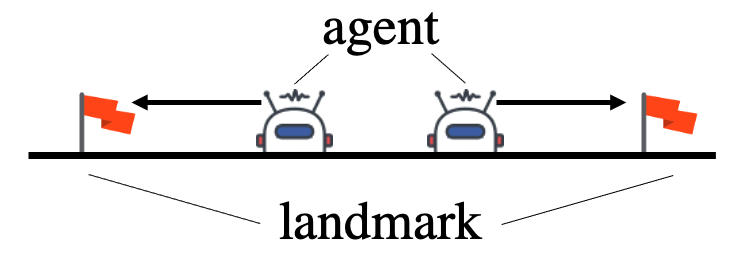}
\vspace{-.2in}
\caption{The Spread environment.}
\label{fig:spread_env}
\vspace{-.2in}
\end{wrapfigure}
We design a Spread environment as shown in Figure \ref{fig:spread_env} which involves $n$ agents and $n$ landmarks ($n \geq 1$) with a 1-dimensional action space to demonstrate the problem and reveal interesting findings.
For the multi-agent setting in the Spread task, $n$ agents need to learn how to cooperate to cover all the landmarks and avoid colliding with each other or arriving at the same landmark. Therefore, it is important for agents to carefully coordinate their actions. 
The experimental setup is the same as in Section \ref{sec:exp}.

\begin{figure}[!h]
\centering
\subfloat[]{\includegraphics[width=0.27\linewidth]{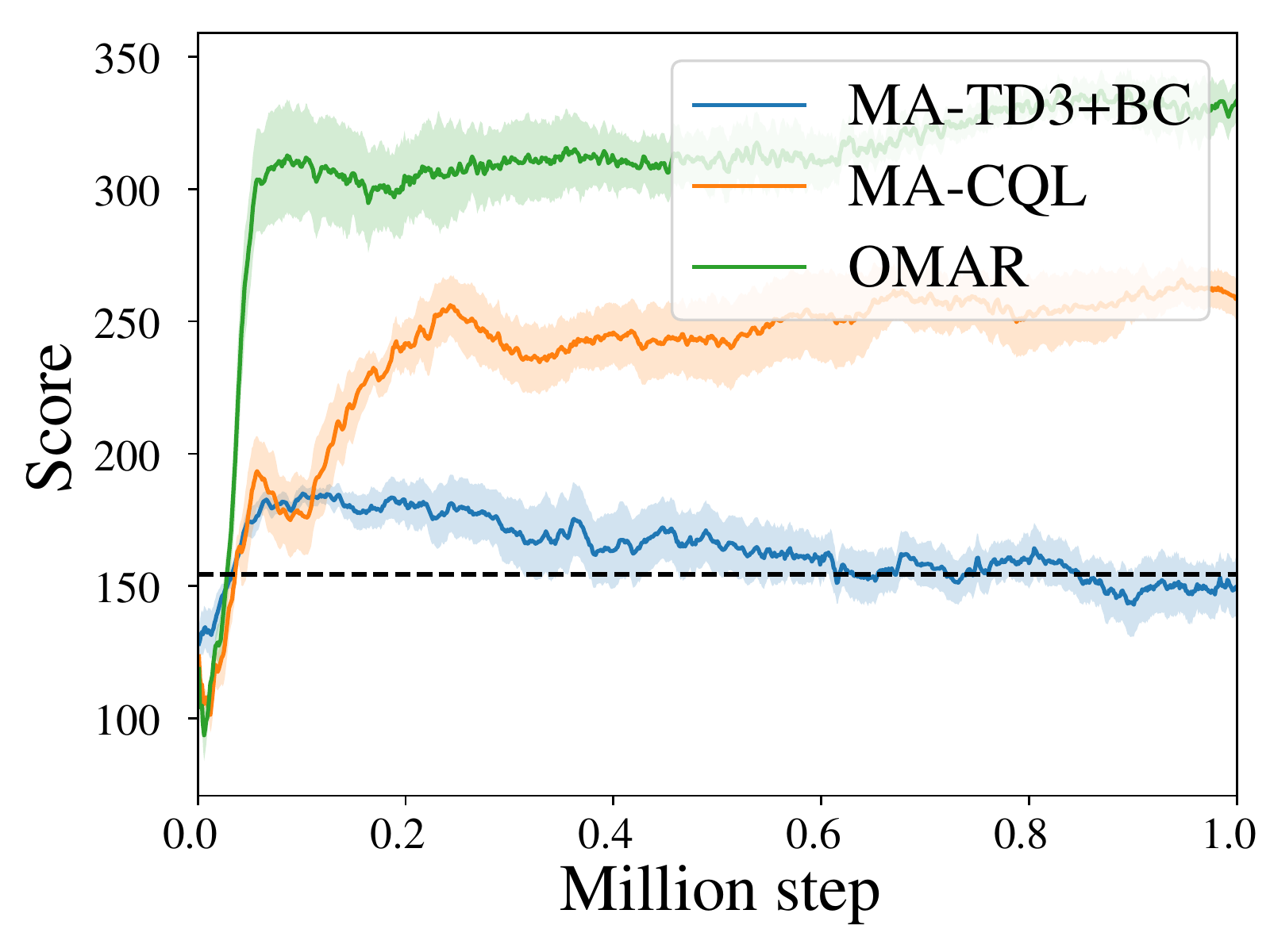}}
\subfloat[]{\includegraphics[width=0.27\linewidth]{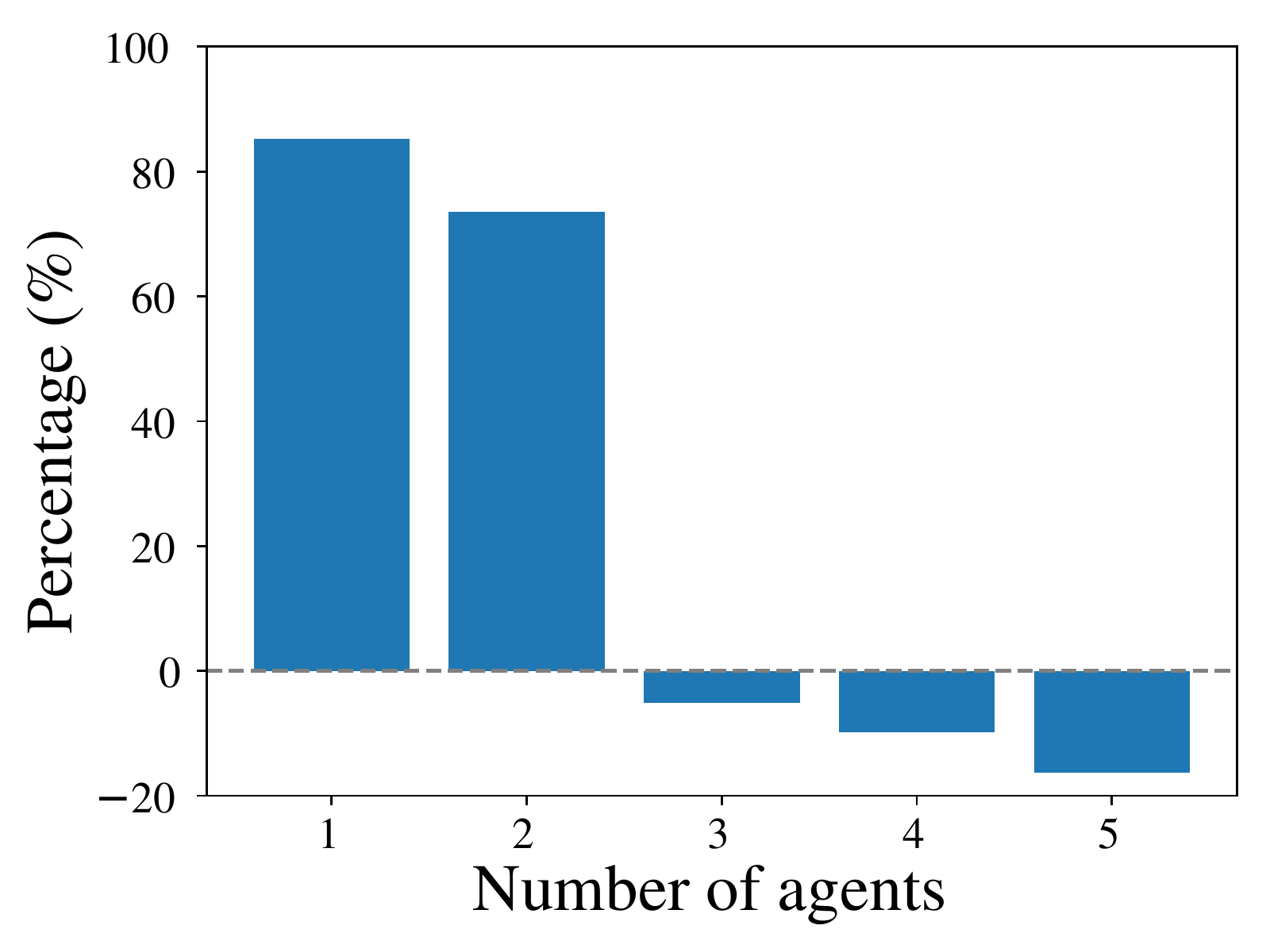}}
\subfloat[]{\includegraphics[width=0.27\linewidth]{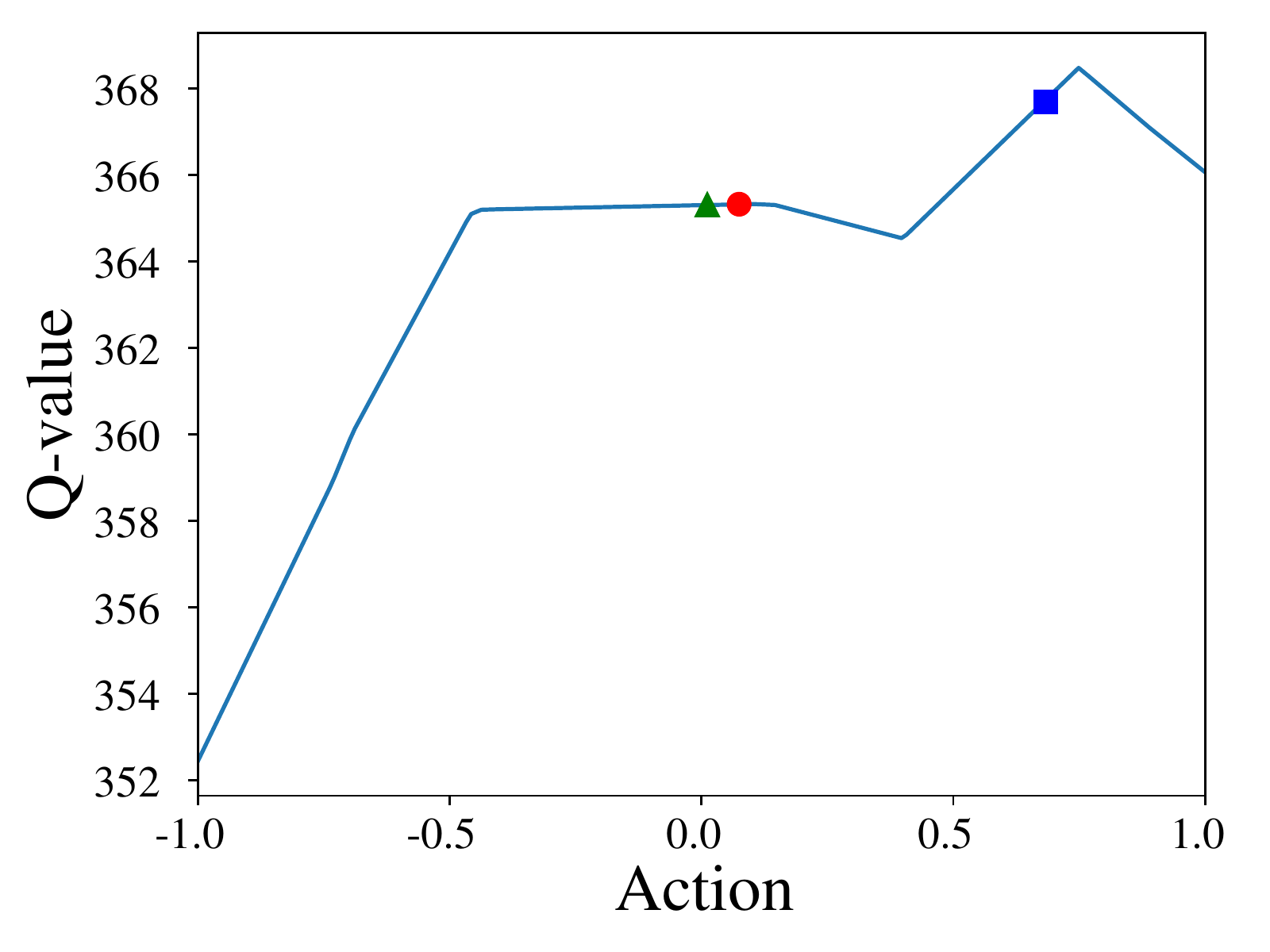}}
\caption{Analysis of MA-TD3+BC, MA-CQL, and OMAR in the medium-replay dataset from Spread. (a) Performance. (b) Performance improvement percentage of MA-CQL over the behavior policy with a varying number of agents. (c) Visualization of the Q-function landscape. The red circle represents the predicted action from the agent using MA-CQL. The green triangle and blue square represent the predicted action from the updated policy of MA-CQL and OMAR.}
\label{fig:simple_navi}
\end{figure}

Figure \ref{fig:simple_navi}(a) demonstrates the performance comparison of the multi-agent version of TD3+BC~\citep{fujimoto2021minimalist}, CQL~\citep{kumar2020conservative}, and OMAR based on ITD3 in the medium-replay dataset from the two-agent Spread environment. 
As MA-TD3+BC is based on policy regularization that compels the learned policy to stay close to the behavior policy, its performance largely depends on the quality of the dataset. Moreover, it can be detrimental to regularize policies to be close to the dataset in multi-agent settings due to decentralized training and the resulting partial observations.
MA-CQL instead learns a lower-bound Q-function to prevent overestimations with additional terms to push down Q-values sampled from a policy while pushing up Q-values for state-action pairs in the dataset. As shown, MA-CQL significantly outperforms MA-TD3+BC in medium-play with more diverse data distribution.

However, despite the effectiveness of MA-CQL when exposed to suboptimal trajectories, we surprisingly find that its performance degrades significantly as there are more agents in the cooperative game. This is shown in Figure \ref{fig:simple_navi}(b), which demonstrates the performance improvement percentage of MA-CQL over the behavior policy with an increasing number of agents from one to five. 

Towards mitigating the performance degradation, we identify a key issue in MA-CQL that solely regularizing the critic is insufficient for multiple agents to learn good policies for coordination. 
In Figure \ref{fig:simple_navi}(c), we visualize the Q-function landscape of MA-CQL during training for an agent in a timestep, with the red circle corresponding to the predicted action from the actor. The green triangle represents the action predicted from the actor after the training step, where the policy gets stuck in a bad local optimum.
The first-order policy gradient method is prone to local optima~\citep{dauphin2014identifying,ahmed2019understanding}, and
we find that the agent can fail to globally leverage the conservative value function well thus leading to suboptimal, uncoordinated learning behavior. 
Note that the problem is severely exacerbated in the offline multi-agent setting due to the exponentially sized joint action space in the number of agents~\citep{yang2021believe}. In addition, it usually requires \emph{each} of the agent to learn a good policy for coordination to solve the task, and the suboptimal policy by any agent could result in uncoordinated global failure.
Note that we also investigate MA-CQL in a non-cooperative version of the Spread task in Appendix \ref{app:cooperation}, whose performance does not degrade with an increasing number of agents. This is because the task does not require careful coordination among agents' policies, which highlights the particularly detrimental effect of this problem in offline MARL.

Increasing the learning rate or the number of updates for actors in MA-CQL does not resolve the problem, where results can be found in Appendix \ref{sec:detailed_mot}.
As a result, to solve this critical challenge, it requires a novel solution instead of blindly tuning hyperparameters. 

\subsection{Offline MARL with Actor Rectification}
Our key observations above identify a critical challenge in offline MARL that
policy gradient improvements are prone to local optima given a bad value function landscape, since the cooperative task requires careful coordination and is sensitive to suboptimal actions.

Zeroth-order optimization methods, \textit{e.g.}, evolution strategies~\citep{rubinstein2013cross,such2017deep,conti2017improving,salimans2017evolution,mania2018simple}, offer an alternative for policy optimization that is also robust to local optima~\citep{rubinstein2013cross}.
It has shown a welcoming avenue towards using zeroth-order methods for policy optimization in the parameter space that improves exploration in the online RL setting~\citep{pourchot2018cem}.

\begin{algorithm}[!h]
\caption{\underline{O}ffline \underline{M}ulti-Agent Reinforcement Learning with \underline{A}ctor \underline{R}ectification (OMAR).}
\label{algo}
\label{alg:RP}
\begin{algorithmic}[1]
\State Initialize $Q$-networks $Q_i^1$, $Q_i^2$, policy networks $\pi_i$ with random parameters $\theta_1^i$, $\theta_2^i$, $\phi_i$, and target networks with $\bar{\theta}_i^1 \leftarrow \theta_i^1$, $\bar{\theta}_i^2 \leftarrow \theta_i^2$, $\bar{\phi}_i \leftarrow \phi_i$ for each agent $i \in [1, N]$
\For {training step $t=1$ to $T$}
\For {agent $i=1$ to $N$}
\State Sample a random minibatch of $S$ samples $(o_i, a_i, r_i, o'_i)$ from $\mathcal{B}$
\State Set $y = r_i + \gamma \min_{j=1,2} \left( \bar{Q}_i^j(o'_i, \pi_i(o'_i + \epsilon)) \right)$ 
\State Update critics $\theta_i$ to minimize Eq. (\ref{eq:cql}) 
\State {Initialize $\mathcal{N}({\mu}_i, {\sigma}_i)$}
\For {iteration $j=1$ to $J$}
\State Draw a population with $K$ individuals $\mathcal{\hat{A}}_i=\{\hat{a}_i^k \sim \mathcal{N}({\mu}_i, {\sigma}_i)\}_{k=1}^K$
\State Estimate $Q$-values for $K$ individuals in the population $\{Q_{i}^1(o_i, \hat{a}_i^k)\}_{k=1}^K$ 
\State Update ${\mu}_i$ and ${\sigma}_i$ according to Eq. (\ref{eq:dist_update})
\EndFor
\State Obtain the picked candidate action $\hat{a}_i=\arg\max_{\hat{a}_i \in \mathcal{\hat{A}}_i \cup \pi_i(o_i)} Q_i^1(o_i,\hat{a}_i)$
\State Update the actor $\phi_i$ to minimize Eq. (\ref{eq:rac})
\State Update target networks: $\bar{\theta}_i^j \leftarrow \rho \theta_i^j + (1-\rho) \bar{\theta}_i^j$ and $\bar{\phi}_i \leftarrow \rho \phi_i + (1-\rho) \bar{\phi}_i$
\EndFor
\EndFor
\end{algorithmic}
\label{algo:rac}
\end{algorithm}

Based on this inspiration, we propose \underline{O}ffline \underline{M}ulti-Agent Reinforcement Learning with \underline{A}ctor \underline{R}ectification (OMAR), which incorporates sampled actions based on Q-values to rectify the actor so that it can escape from bad 
local optima.
For simplicity of presentation, we demonstrate our method based on the decentralized training paradigm introduced in Section \ref{sec:maac}, which can also be applied to centralized critics as shown in Appendix \ref{sec:matd3}.
Specifically, we propose the following policy objective by introducing a regularizer:
\begin{align}
\mathbb{E}_{\mathcal{D}_i} \left [(1-\tau) Q_i(o_i, \pi_i(o_i)) - \tau \left( \pi_i(o_i) - \hat{a}_i \right)^2 \right]
\label{eq:rac}
\end{align}
where $\hat{a}_i$ is the action provided by the zeroth-order optimizer and $\tau \in [0, 1]$ denotes the coefficient.
Note that TD3+BC~\citep{fujimoto2021minimalist} can be interpreted as using the seen action in the dataset for $\hat{a}_i$. The distinction between 
\textit{optimized} and \textit{seen} actions enables OMAR to perform well even if the dataset quality is from mediocre to low.

We propose our sampling mechanism motivated by 
the cross-entropy method (CEM)~\citep{rubinstein2013cross}, which has shown great potential in RL~\citep{lim2018actor}.
However, CEM does not scale to tasks with high-dimensional space well~\citep{nagabandi2020deep}.
We instead propose to sample actions in a softer way motivated by~\citet{williams2015model,lowrey2018plan}.
Specifically, we sample actions according to an iteratively refined Gaussian distribution $\mathcal{N}(\mu_i, \sigma_i)$. At each iteration $j$, we draw $K$ candidate actions by $a_i^j\sim \mathcal{N}(\mu_i^j, \sigma_i^j)$ and evaluate all their Q-values. The mean and standard deviation of the sampling distribution is updated and refined according to Eq. (\ref{eq:dist_update}), which produces a softer update and leverages more samples in the update~\citep{nagabandi2020deep}. Our sampling mechanism indeed outperforms CEM and random as illustrated in Section \ref{sec:abl}. Our method is built upon CQL, which lower-bounds the true Q-function to largely reduce overestimation. The resulting OMAR method is shown in Algorithm \ref{algo:rac}. 
\begin{equation}
\mu_i^{j+1} = \frac{\sum_{k=1}^K \exp(\beta Q_i^k) a_i^k}{\sum_{m=1}^K \exp(\beta Q_i^m)}, \quad \sigma_i^{j+1} = \sqrt{\sum_{k=1}^K \left(a_i^k-\mu_i^j \right)^2}.
\label{eq:dist_update}
\end{equation}

Next, we theoretically justify that OMAR provides a safe policy improvement guarantee. Let $J(\pi_i)$ denote the discounted return of a policy $\pi_i$ in the empirical MDP $\hat{M}_i$ which is induced by transitions in the dataset $\mathcal{D}_i$, i.e., $\hat{M}_i=\{(o_i, a_i, r_i, o_i^{\prime}) \in \mathcal{D}_i\}$. In Theorem \ref{thm:safe}, we give a lower bound
on the difference between the policy performance of OMAR over the empirical behavior policy 
$\hat{\pi}_{\beta_i}$ in the empirical MDP $\hat{M}_i$. The proof is in Appendix \ref{app:proof}. 
\begin{theorem}
Let $\pi_i^*$ denote the policy obtained by optimizing Eq. (\ref{eq:rac}), $D(\pi_i,\hat{\pi}_{\beta_i})(o_i) = \frac{1-\hat{\pi}_{\beta_i}(\pi_i(o_i)|o_i)}{\hat{\pi}_{\beta_i}(\pi_i(o_i)|o_i)}$, and $d^{\pi_i}(o_i)$ denote the marginal discounted distribution of observations of policy $\pi_i$. Then, we have that $J(\pi_i^*) - J(\hat{\pi}_{\beta_i}) \geq  \frac{\alpha}{1-\gamma} \mathbb{E}_{o_i\sim d^{\pi_i^*}(o_i)}\left[D(\pi_i^*,\hat{\pi}_{\beta_i})(o_i)\right]$
$+ \frac{\tau}{1-\tau}\mathbb{E}_{o_i\sim d^{\pi_i^*}(o_i)} \left[ (\pi_i^*(o_i)-\hat{a}_i)^2 \right]$\\
$- \frac{\tau}{1-\tau}\mathbb{E}_{o_i\sim d^{\hat{\pi}_{\beta_i}}(o_i),a_i\sim\hat{\pi}_{\beta_i}} \left[ (a_i-\hat{a}_i)^2 \right].$
\label{thm:safe}
\end{theorem}
\paragraph{Remark.} From Theorem \ref{thm:safe}, the first term on the right-hand side is non-negative, and the difference between the second and third terms is the difference between two expected distances. The former corresponds to the gap between the action from our zeroth-order optimizer $\hat{a}_i$ and the optimal action $\pi_{i}^{*}(o_i)$.
The latter corresponds to the gap between $\hat{a}_i$ and the action from the behavior policy. Since both terms can be bounded and controlled, we find that OMAR gives a safe policy improvement guarantee over $\hat{\pi}_{\beta_i}$.

\subsubsection{The Effect of OMAR in the Spread Task}
Now, we investigate whether OMAR can successfully address the identified problem using the Spread environment as an example. We further analyze its effect in offline/online, multi-agent/single-agent settings for a better understanding of the potential of our method.

\begin{figure}[!h]
\centering
\includegraphics[width=0.35\linewidth]{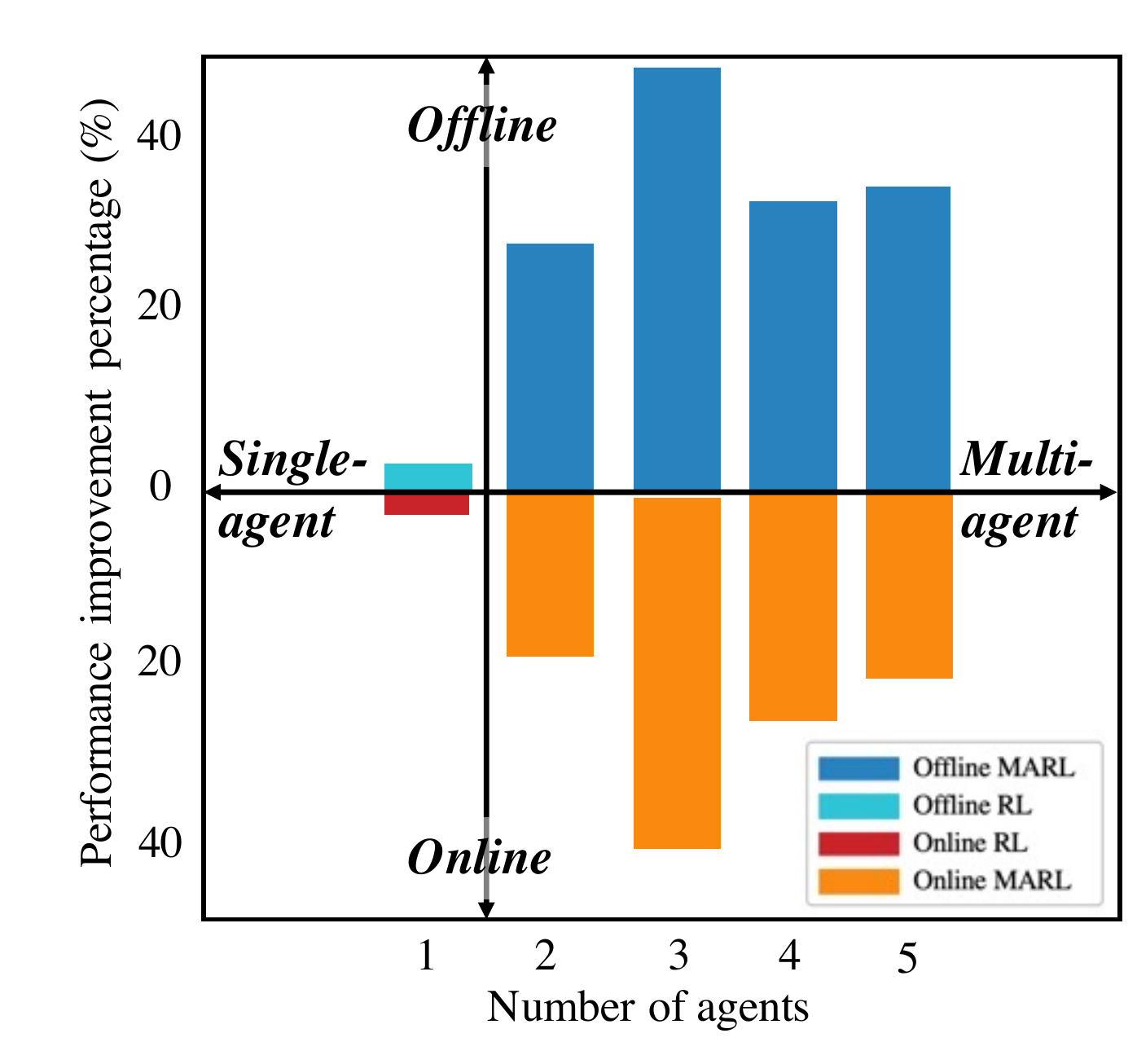}
\caption{Performance improvement percentage of our method over MA-CQL in the offline (upper part) setting and MATD3 in the online setting (lower part) with a varying number of agents in the Spreak task. The first to fourth quadrants correspond to the offline MARL, offline RL, online RL, and online MARL settings.}
\label{fig:online_discuss}
\end{figure}

\paragraph{Can OMAR address the identified problem?}
In Figure~\ref{fig:simple_navi}(c), the blue square corresponds to the action from the updated actor using OMAR according to Eq. (\ref{eq:rac}). In contrast to the policy update in MA-CQL, OMAR can better leverage the global information in the critic and help the actor to escape from the bad local optimum.
Figure \ref{fig:simple_navi}(a) further validates that OMAR significantly improves MA-CQL in terms of both performance and efficiency.
The upper part in Figure \ref{fig:online_discuss} shows the performance improvement percentage of OMAR over MA-CQL (y-axis) with a varying number of agents (x-axis), where OMAR always outperforms MA-CQL. We also notice that the performance improvement of OMAR over MA-CQL is much more significant in the multi-agent setting in Spread than in the single-agent setting. This echoes with what is discussed above that the problem becomes more critical with more agents, as it requires each of the agents to learn a good policy based on the conservative value function for a successful joint policy for coordination. Otherwise, it can lead to an uncoordinated global failure.

\paragraph{Is OMAR effective in offline/online, multi-agent/single-agent settings?}
We next investigate the effectiveness of OMAR in the following settings corresponding to different quadrants in Figure \ref{fig:online_discuss}: i) offline multi-agent, ii) offline single-agent, iii) online single-agent, iv) online multi-agent. 
For the online setting, we build our method upon MATD3 based on clipped double estimators with our proposed policy objective in Eq. (\ref{eq:rac}). We evaluate the performance improvement percentage of our method over MATD3. 
The results for the online setting are shown in the lower part in Figure \ref{fig:online_discuss}.

As shown in Figure \ref{fig:online_discuss}, our method is generally applicable in all the settings, with a much more significant  performance improvement in the offline setting (upper part) than the online case (lower part).
Intuitively, in the online setting, if the actor has not well exploited the global information in the value function, it can still explore and interact with the environment to collect better experiences for improving the value estimation and provides better guidance for the policy. 
However, in the offline setting, it is much harder for an agent to escape from a bad local optimum and difficult for the actor to best leverage the global information in the conservative critic. 
As expected, we find that the performance gain is the largest in the offline multi-agent domain.  
\begin{table*}[!h]
\caption{Averaged normalized score of OMAR and baselines in multi-agent particle environments.}
\label{tab:mpe}
\centering
\begin{tabular}{cccccccc}
\toprule
\multicolumn{2}{c}{} & MA-ICQ & MA-TD3+BC & MA-CQL & OMAR \\
\midrule
\multirow{3}{*}{\rotatebox[origin=c]{90}{Random}} & Cooperative navigation & $6.3\pm3.5$ & $9.8\pm4.9$ & $24.0\pm9.8$ & $\textbf{34.4}\pm5.3$\\
& Predator-prey & $2.2\pm2.6$ & $5.7\pm3.5$ & $5.0\pm8.2$ & $\textbf{11.1}\pm2.8$ \\
& World & $1.0\pm3.2$ & $2.8\pm5.5$ & $0.6\pm2.0$ & $\textbf{5.9}\pm5.2$ \\
\midrule
\multirow{3}{*}{\rotatebox[origin=c]{90}{\shortstack{Medium\\-replay}}} & Cooperative navigation & $13.6\pm5.7$ & $15.4\pm5.6$ & $20.0\pm8.4$  & $\textbf{37.9}\pm12.3$ \\
& Predator-prey & $34.5\pm27.8$ & $28.7\pm20.9$ & $24.8\pm17.3$ & $\textbf{47.1}\pm15.3$ \\
& World & $12.0\pm9.1$ & $17.4\pm8.1$ & $29.6\pm13.8$ & $\textbf{42.9}\pm19.5$ \\
\midrule
\multirow{3}{*}{\rotatebox[origin=c]{90}{Medium}} & Cooperative navigation & $29.3\pm5.5$ & $29.3\pm4.8$ & $34.1\pm7.2$ & $\textbf{47.9}\pm18.9$ \\
& Predator-prey & $63.3\pm20.0$ & $65.1\pm29.5$ & $61.7\pm23.1$ & $\textbf{66.7}\pm23.2$ \\
& World & $71.9\pm20.0$ & $73.4\pm9.3$ & $58.6\pm11.2$ & $\textbf{74.6}\pm11.5$\\
\midrule
\multirow{3}{*}{\rotatebox[origin=c]{90}{Expert}} & Cooperative navigation & $104.0\pm3.4$ & $108.3\pm3.3$ & $98.2\pm5.2$  & $\textbf{114.9}\pm2.6$ \\
& Predator-prey & $113.0\pm14.4$ & $115.2\pm12.5$ & $93.9\pm14.0$ & $\textbf{116.2}\pm19.8$ \\
& World & $109.5\pm22.8$ & $110.3\pm21.3$ & $71.9\pm28.1$ & $\textbf{110.4}\pm25.7$ \\
\bottomrule
\end{tabular}
\end{table*}

\section{Experiments} \label{sec:exp}
We conduct a series of experiments to study the following key questions: i) How does OMAR compare against state-of-the-art offline RL and MARL methods? ii) What is the effect of critical hyperparameters, our sampling mechanism, and the size of the dataset? iii) Can OMAR scale to the more complex multi-agent locomotion tasks? iv) Can OMAR be applied to the challenging discrete control StarCraft II micromanagement benchmarks? v) Is OMAR compatible in single-agent tasks?
Videos are available at \url{https://sites.google.com/view/omar-videos}.

\subsection{Multi-Agent Particle Environments} \label{sec:mpe}
We first conduct a series of experiments in the widely-adopted multi-agent particle environments~\citep{lowe2017multi} where the agents need to cooperate to solve the task. A detailed description of the cooperative navigation, predator-prey, and world environments is in Appendix \ref{sec:task}.
We construct a variety of datasets according to behavior policies with different qualities based on adding noises to MATD3 to increase diversity following~\citep{fu2020d4rl}. The random dataset is generated by rolling out a randomly initialized policy for 1 million (M) steps. We obtain the medium-replay dataset by recording all samples in the replay buffer during training until the policy reached the medium level of the performance. The medium or expert datasets consist of 1M samples by unrolling a partially-pretrained policy with a medium performance level or a fully-trained policy.

We compare OMAR against state-of-the-art offline RL methods including CQL~\citep{kumar2020conservative} and TD3+BC~\citep{fujimoto2021minimalist}. We also compare it with a recent offline MARL algorithm MA-ICQ~\citep{yang2021believe}. 
We build all the methods on the independent TD3 based on decentralized critics, while we also consider centralized critics based on MATD3 with a detailed evaluation in Appendix \ref{sec:matd3} that achieves similar performance improvement.
Each algorithm is run for five random seeds, and we report the mean performance with standard deviation.
A detailed description of hyperparameters and setup can be found in Appendix \ref{sec:baselines}.

\subsubsection{Performance Comparison}
Table \ref{tab:mpe} summarizes the average normalized scores in different datasets in multi-agent particle environments, where the learning curves are shown in Appendix \ref{app:res}. The normalized score is computed as $100 \times (S - S_{\text{random}})/(S_{\text{expert}} - S_{\text{random}})$ following~\citet{fu2020d4rl}.
As shown, the performance of MA-TD3+BC highly depends on the quality of the dataset.
MA-ICQ is based on only trusting seen state-action pairs in the dataset. As shown, it does not perform well in datasets with more diverse data distribution (random and medium-replay), while generally matching the performance of MA-TD3+BC in datasets with narrower distribution (medium and expert). MA-CQL matches or outperforms MA-TD3+BC in datasets with lower quality except for the expert dataset, as it does not rely on constraining the learning policy to stay close to the behavior policy. 
OMAR significantly outperforms all the baselines and achieves state-of-the-art performance. We attribute the performance gain to the actor rectification scheme that is independent of data quality and improves global optimization. In addition, OMAR does not incur much computation cost and only takes $4.7\%$ more runtime on average compared with that of MA-CQL. 

\subsubsection{Ablation Study} \label{sec:abl}
We now investigate how sensitive OMAR is to key hyperparameters including the regularization coefficient $\tau$ and the effect of the sampling mechanism. We also analyze the effect of the size of the dataset in Appendix \ref{app:size_ab}.

\begin{figure}[!h]
\centering
\subfloat[Random.]{\includegraphics[width=0.25\linewidth]{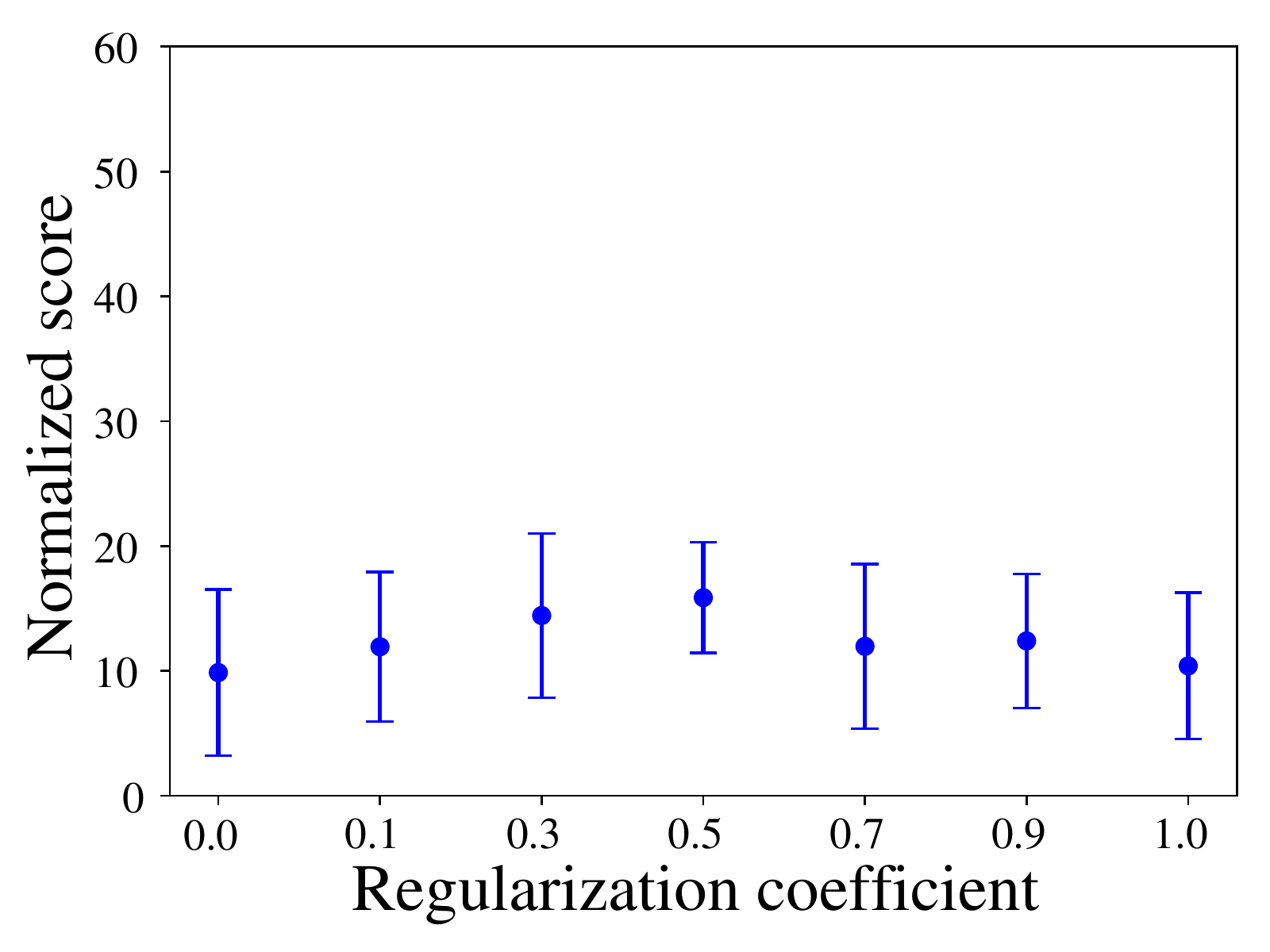}}
\subfloat[Med-replay.]{\includegraphics[width=0.25\linewidth]{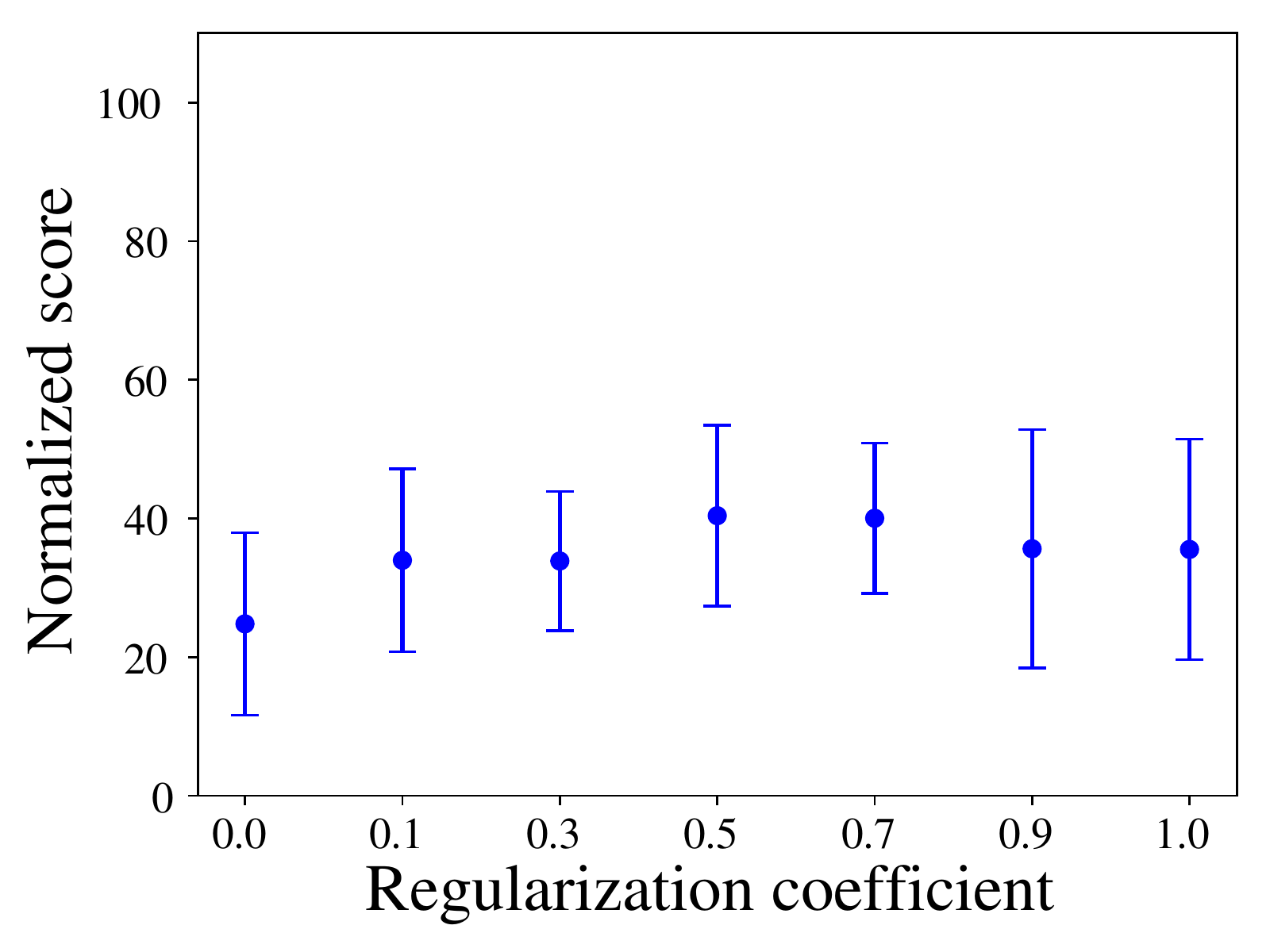}}
\subfloat[Medium.]{\includegraphics[width=0.25\linewidth]{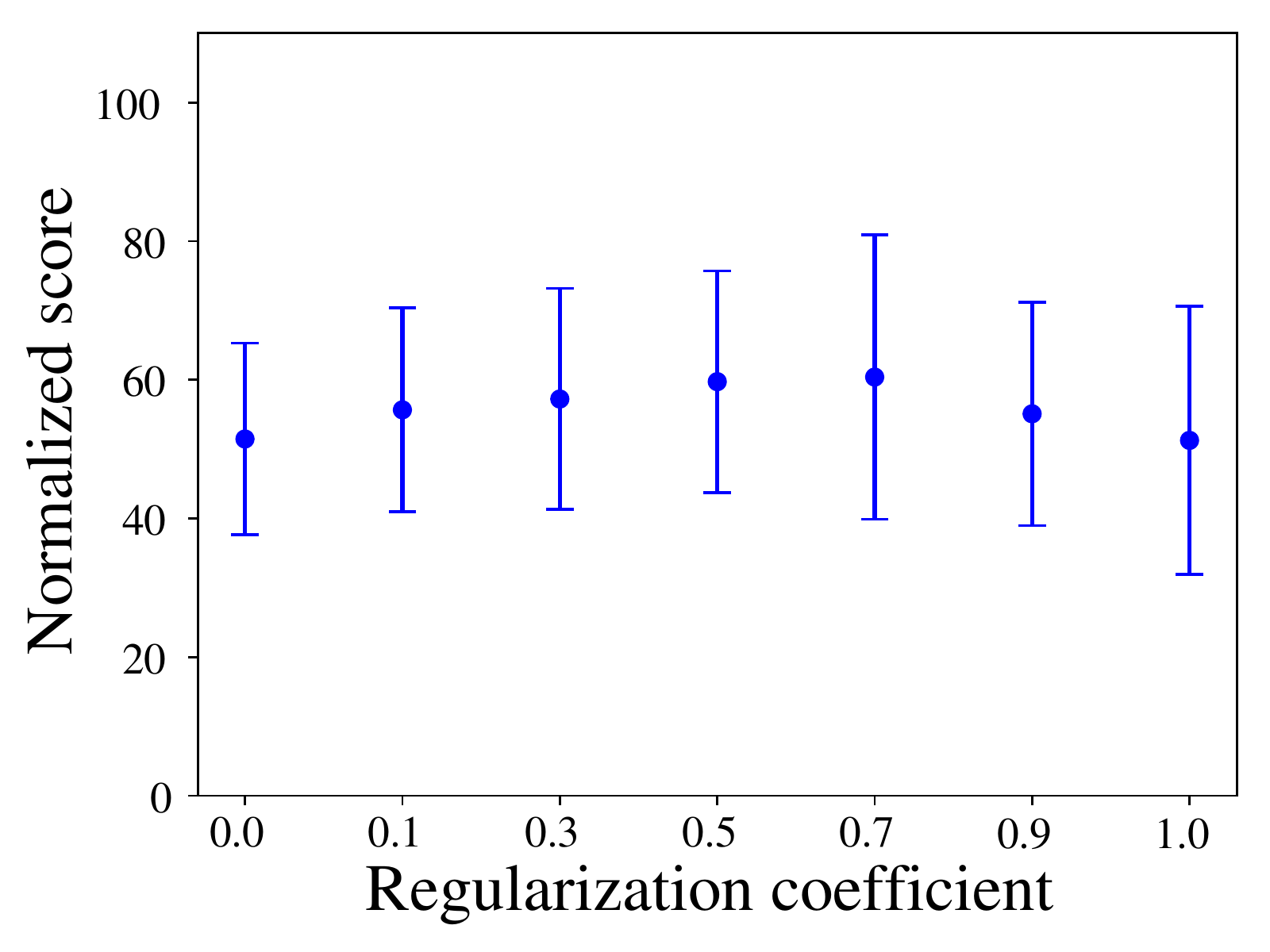}}
\subfloat[Expert.]{\includegraphics[width=0.25\linewidth]{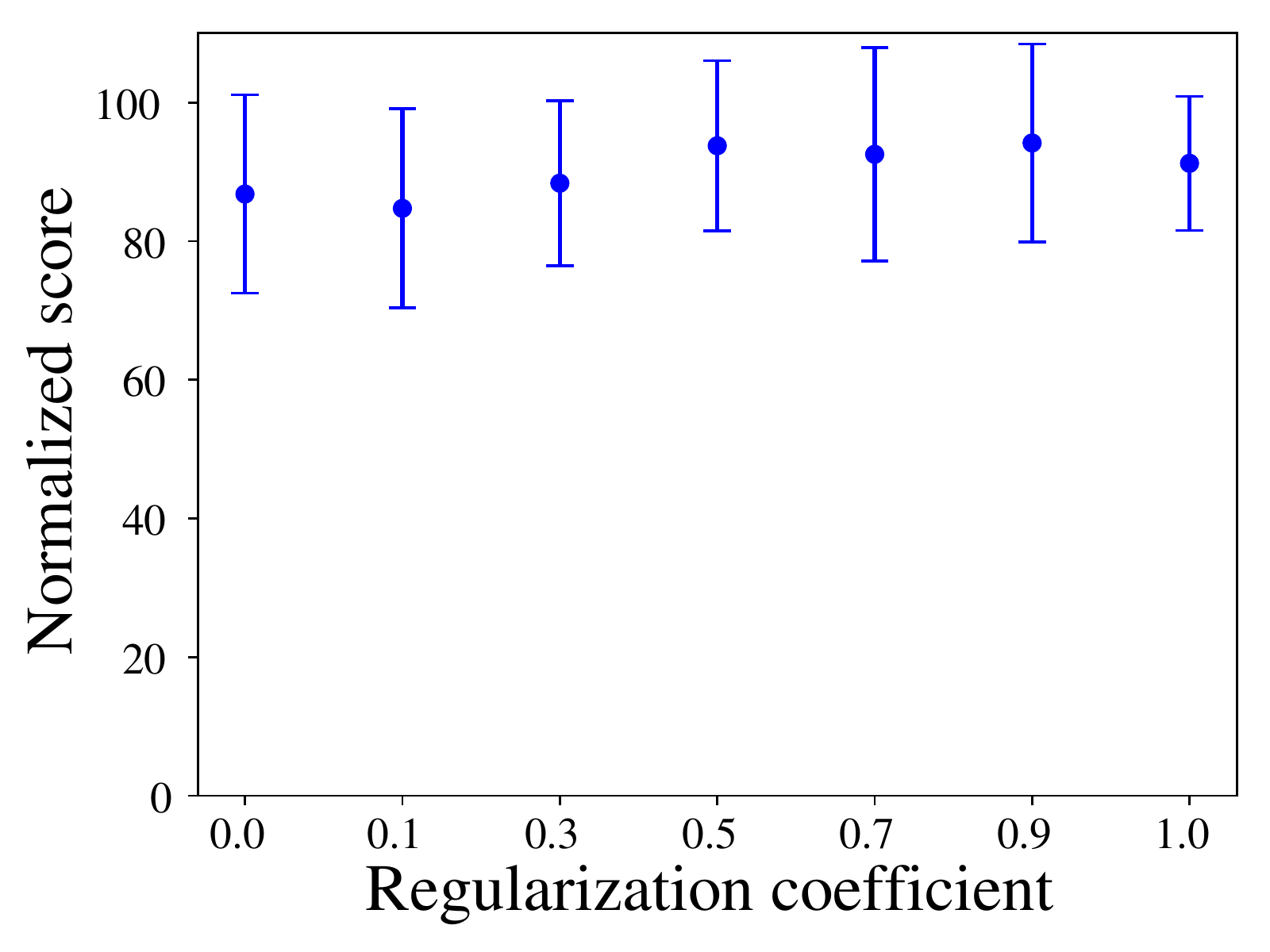}}
\caption{Ablation study on the regularization coefficient in different types of datasets.}
\label{fig:ablation1}
\end{figure}

\paragraph{The effect of the regularization coefficient.} Figure \ref{fig:ablation1} shows the averaged normalized score of OMAR over different tasks with different values of the regularization coefficient $\tau$ in each kind of dataset. As shown, OMAR is sensitive to this hyperparameter, which controls the exploitation level of the critic. We find the best value of $\tau$ is neither close to $1$ nor $0$, showing that it is the combination of both policy gradients and the actor rectification that performs well. We also notice that the optimal value of $\tau$ is smaller for datasets with lower quality and more diverse data distribution including random and medium-replay, but larger for medium and expert datasets. In addition, the performance of OMAR with all values of $\tau$ matches or outperforms that of MA-CQL. 
Note that this is the only hyperparameter that needs to be tuned in OMAR beyond MA-CQL. 

\begin{figure}[!h]
\centering
\subfloat[Iterations.]{\includegraphics[width=0.25\linewidth]{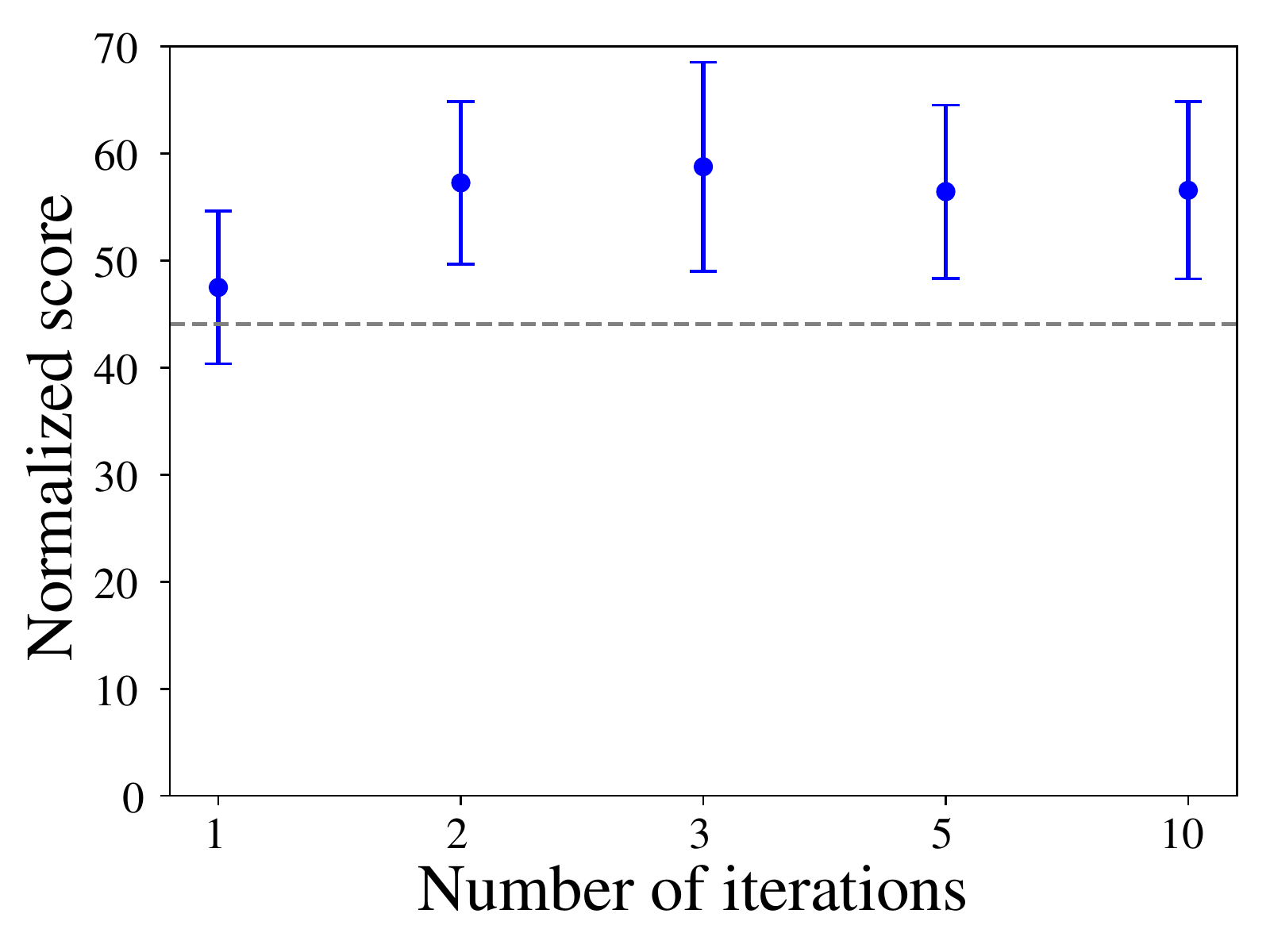}}
\subfloat[Samples.]{\includegraphics[width=0.25\linewidth]{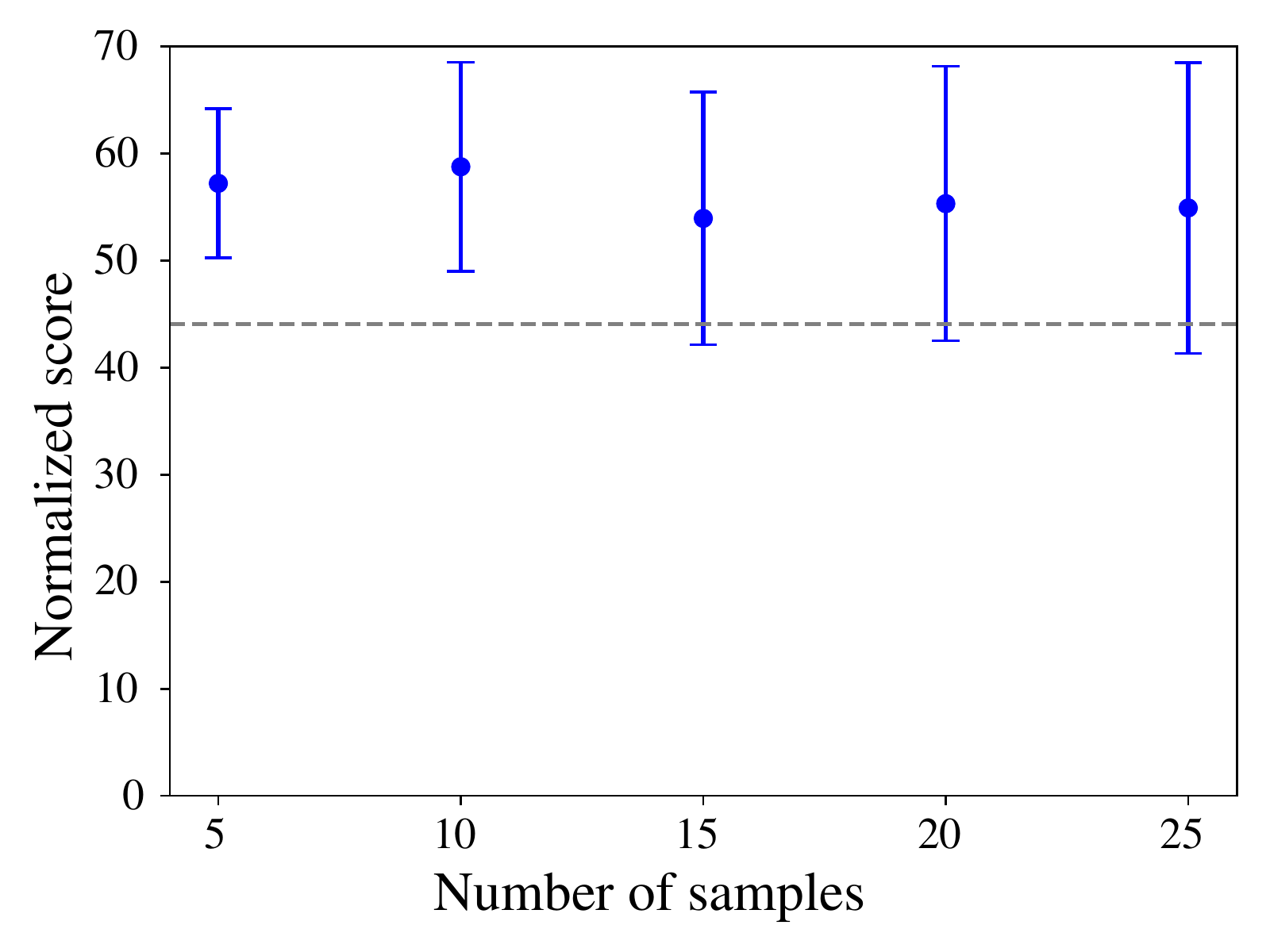}}
\subfloat[Mean.]{\includegraphics[width=0.25\linewidth]{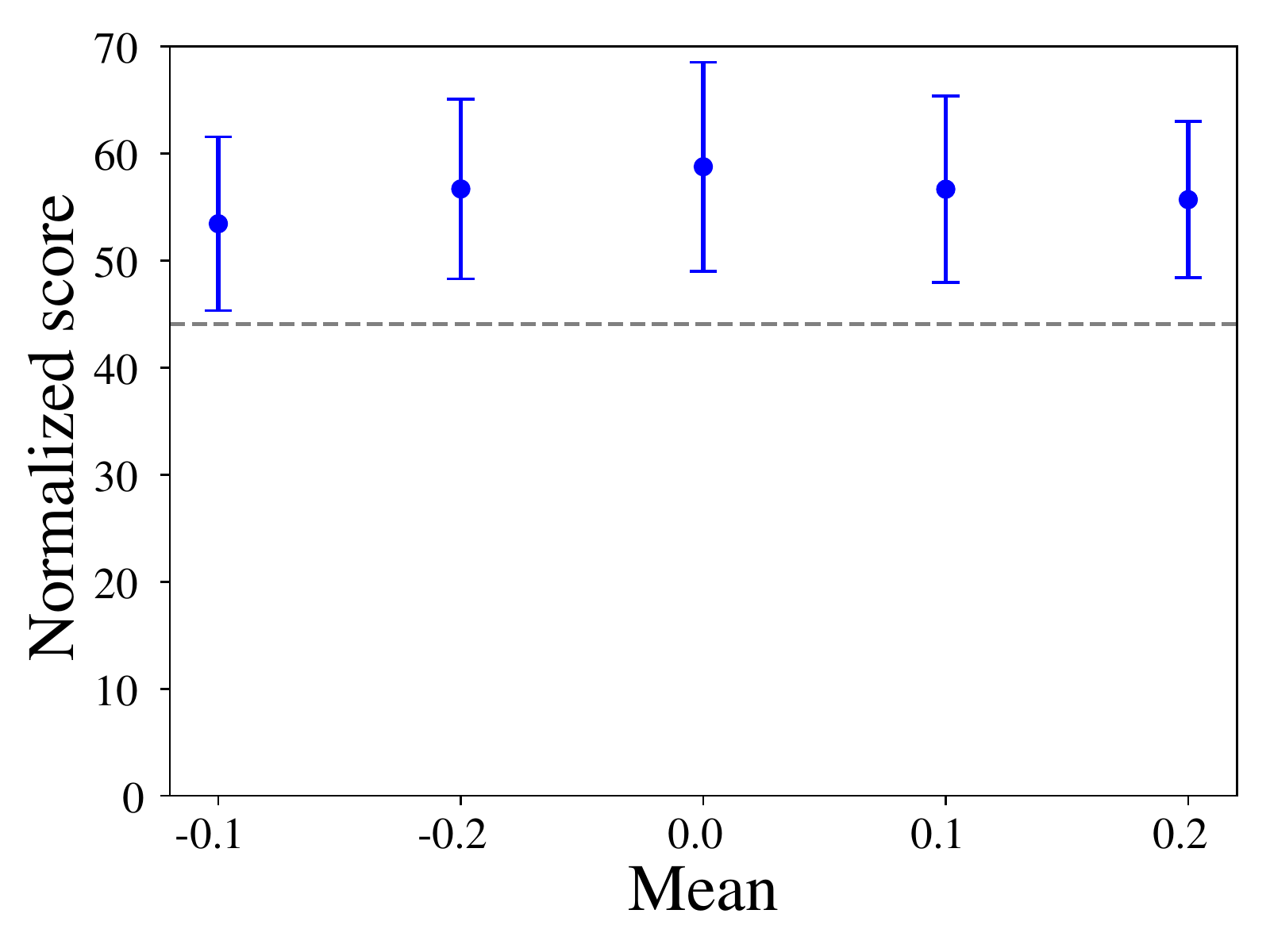}}
\subfloat[Std.]{\includegraphics[width=0.25\linewidth]{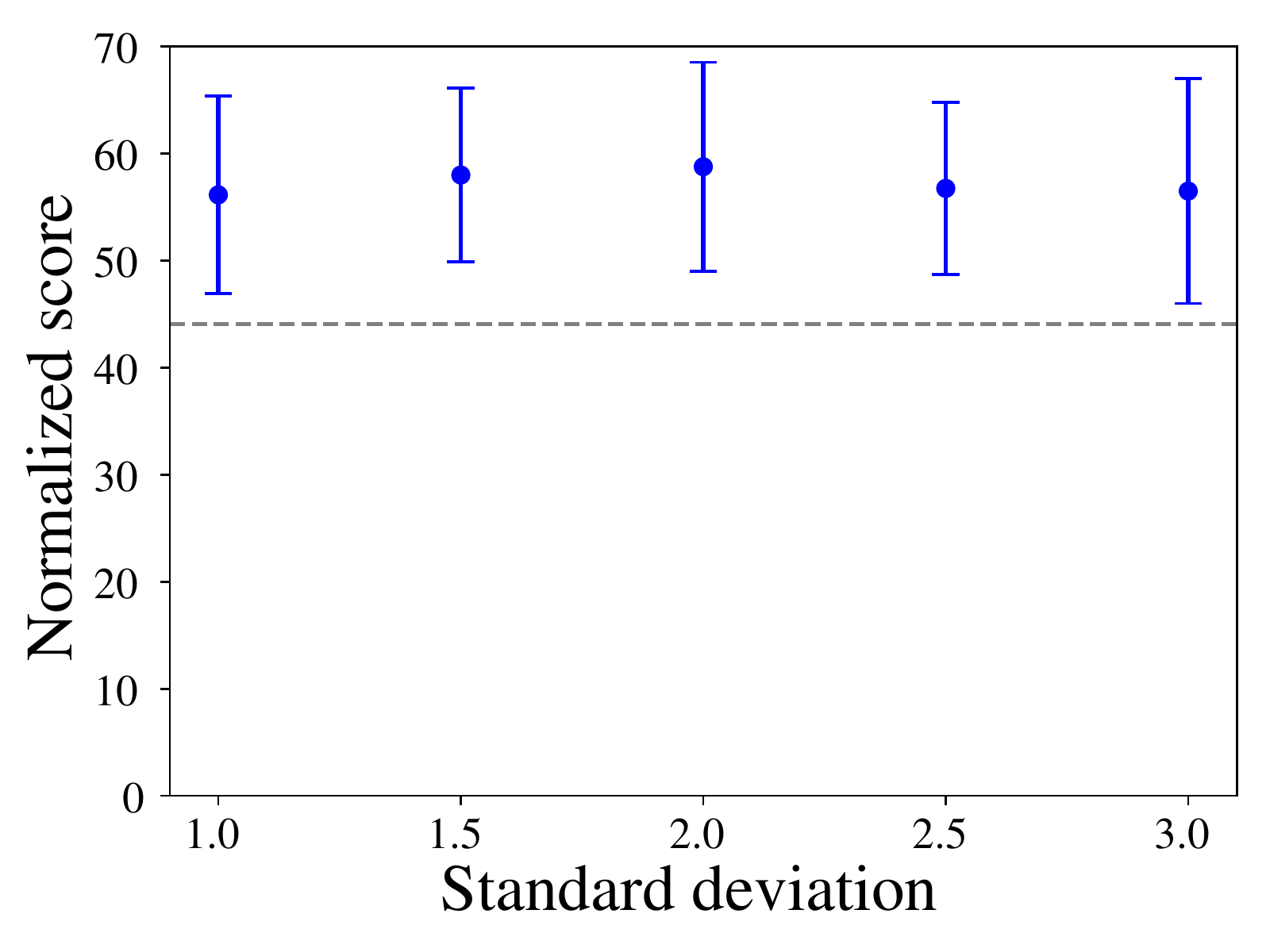}}
\caption{Ablation study on key hyperparameters in the sampling mechanism averaged over different types of datasets.}
\label{fig:ablation2}
\end{figure}

\paragraph{The effect of key hyperparameters in the sampling mechanism.} Core hyperparameters for our sampling scheme involve the number of iterations, the number of sampled actions, and the initial mean and standard deviation of the Gaussian distribution. Figure \ref{fig:ablation2} shows the performance comparison of OMAR with different values of these hyperparameters in the cooperative navigation task, where the grey dotted line corresponds to the normalized score of MA-CQL. As shown, our sampling mechanism is not sensitive to these hyperparameters, and we therefore fix them for all types of the tasks to be the same set with the best performance.

\paragraph{The effect of the sampling mechanism.} We now analyze the effect of the zeroth-order optimizer in OMAR, and compare our sampling scheme against random sampling and the cross-entropy method (CEM)~\citep{de2005tutorial} in the cooperative navigation task. As shown in Table \ref{tab:ablation_sampling}, our sampling mechanism significantly outperforms CEM and random sampling in all types of datasets with different qualities. It enjoys a larger margin in datasets with lower quality including random and medium-replay. 
This is because the proposed sampling mechanism incorporates more samples into the distribution in a softer way, which updates in a more effective way.

\begin{table}[!h]
\caption{Ablation study of OMAR with different sampling mechanisms in different types of datasets.}
\label{tab:ablation_sampling}
\centering
\begin{tabular}{ccccc}
\toprule
 ~ & OMAR (random) & OMAR (CEM) & OMAR \\
\midrule
Random & $24.3\pm7.0$ & $25.8\pm7.3$ & $\textbf{34.4}\pm5.3$ \\ 
Med-rep & $23.5\pm5.3$ & $32.6\pm5.1$ & $\textbf{37.9}\pm5.3$ \\ 
Medium & $41.2\pm11.1$ & $45.0\pm13.3$ & $\textbf{47.9}\pm18.9$ \\ 
Expert & $101.0\pm5.2$ & $106.4\pm13.8$ & $\textbf{114.9}\pm2.6$ \\
\bottomrule
\end{tabular}
\end{table}

\subsection{Multi-Agent MuJoCo}
In this section, we investigate whether OMAR can scale to the more complex continuous control multi-agent task. 
We consider the multi-agent HalfCheetah task from the multi-agent MuJoCo environment~\citep{peng2020facmac}, which extends the high-dimensional MuJoCo locomotion tasks in the single-agent setting to the multi-agent case. In this environment, agents control different parts of joints of the robot as shown in Appendix \ref{sec:task}. These agents need to cooperate to make the robot run forward by coordinating their actions.
Different types of datasets are constructed following the same way as in Section \ref{sec:mpe}.

Table \ref{tab:mamujoco} summarizes the average normalized scores in each kind of dataset in multi-agent HalfCheetah. As shown, OMAR significantly outperforms baseline methods in random, medium-replay, and medium datasets, and matches the performance of MA-TD3+BC in expert, demonstrating its effectiveness to scale to more complex control tasks. It is also worth noting that the performance of MA-TD3+BC depends on the quality of the data, which underperforms OMAR in other types of dataset except for expert.

\begin{table}[!h]
\caption{Average normalized score of different methods in multi-agent HalfCheetah.}
\label{tab:mamujoco}
\centering
\begin{tabular}{ccccc}
\toprule
& ICQ & TD3+BC & CQL & OMAR \\
\midrule
Random & $7.4\pm0.0$ & $7.4\pm0.0$ & $7.4\pm0.0$ & $\textbf{13.5}\pm7.0$ \\
Med-rep & $35.6\pm2.7$ & $27.1\pm5.5$ & $41.2\pm10.1$ & $\textbf{57.7}\pm5.1$ \\
Medium & $73.6\pm5.0$ & $75.5\pm3.7$ & $50.4\pm10.8$ & $\textbf{80.4}\pm10.2$ \\
Expert & $110.6 \pm 3.3$ & $\textbf{114.4}\pm3.8$ & $64.2\pm24.9$ & $\textbf{113.5}\pm4.3$ \\
\bottomrule
\end{tabular}
\end{table}

\subsection{StarCraft II Micromanagement Benchmark}
In this section, we further study the effectiveness of OMAR in larger-scale tasks based on the StarCraft II micromanagement benchmark~\citep{samvelyan2019starcraft} on maps with an increasing number of agents and difficulties including 2s3z, 3s5z, 1c3s5z, and 2c\_vs\_64zg. We compare the most competitive MA-CQL method and OMAR based on the evaluation protocol in \citet{kumar2020conservative,agarwal2020optimistic,gulcehre2020rl}, where datasets are constructed following \citet{agarwal2020optimistic,gulcehre2020rl} by recording samples observed during training. Each dataset consists of 1M samples.
We use the Gumbel-Softmax
reparameterization method~\citep{jang2016categorical} to generate discrete actions for MATD3 since it requires differentiable policies~\citep{lowe2017multi,iqbal2019actor,peng2020facmac}. % as detailed in Appendix \ref{sec:gumbel}.
A detailed description of the tasks and implementation details can be found in Appendix \ref{app:exp}.

Figure \ref{fig:smac} demonstrates the comparison results in test win rates. As shown, OMAR significantly outperforms MA-CQL in performance and learning efficiency. The average performance gain of OMAR compared to MA-CQL is $76.7\%$ in all tested maps, showing that OMAR is also effective in the challenging discrete control StarCraft II micromanagement tasks.
\begin{figure}[!h]
\centering
\includegraphics[width=0.24\linewidth]{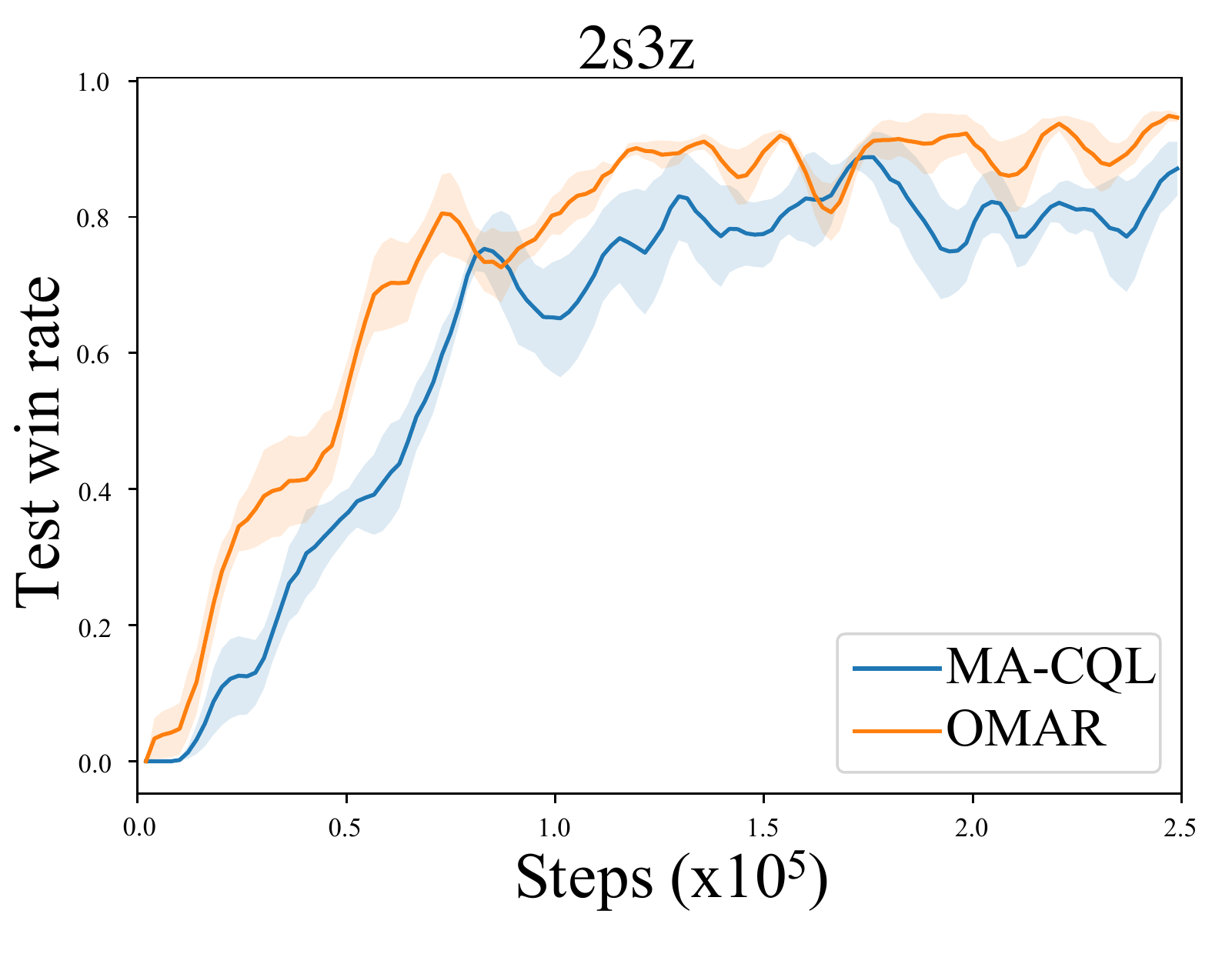}
\includegraphics[width=0.24\linewidth]{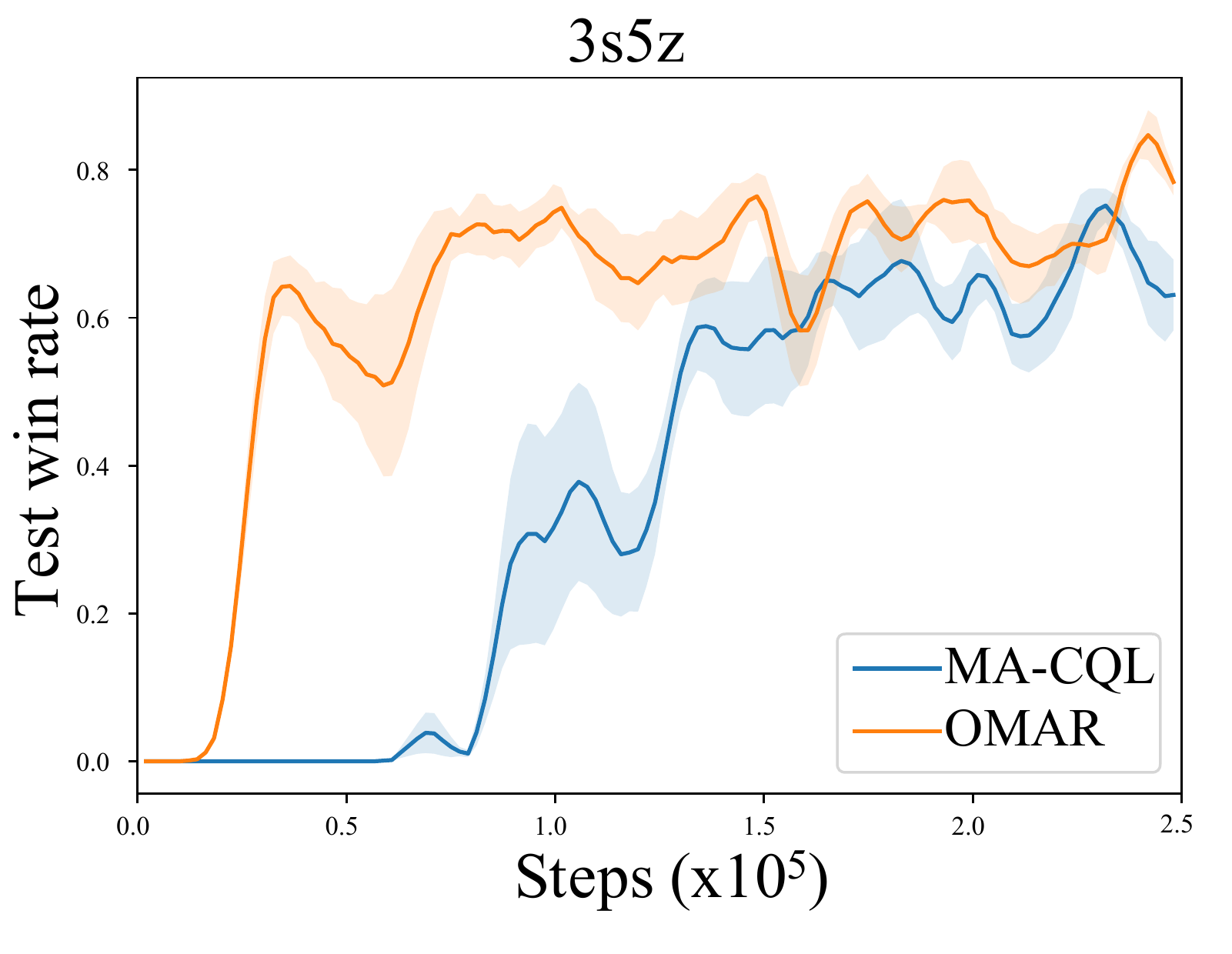}
\includegraphics[width=0.24\linewidth]{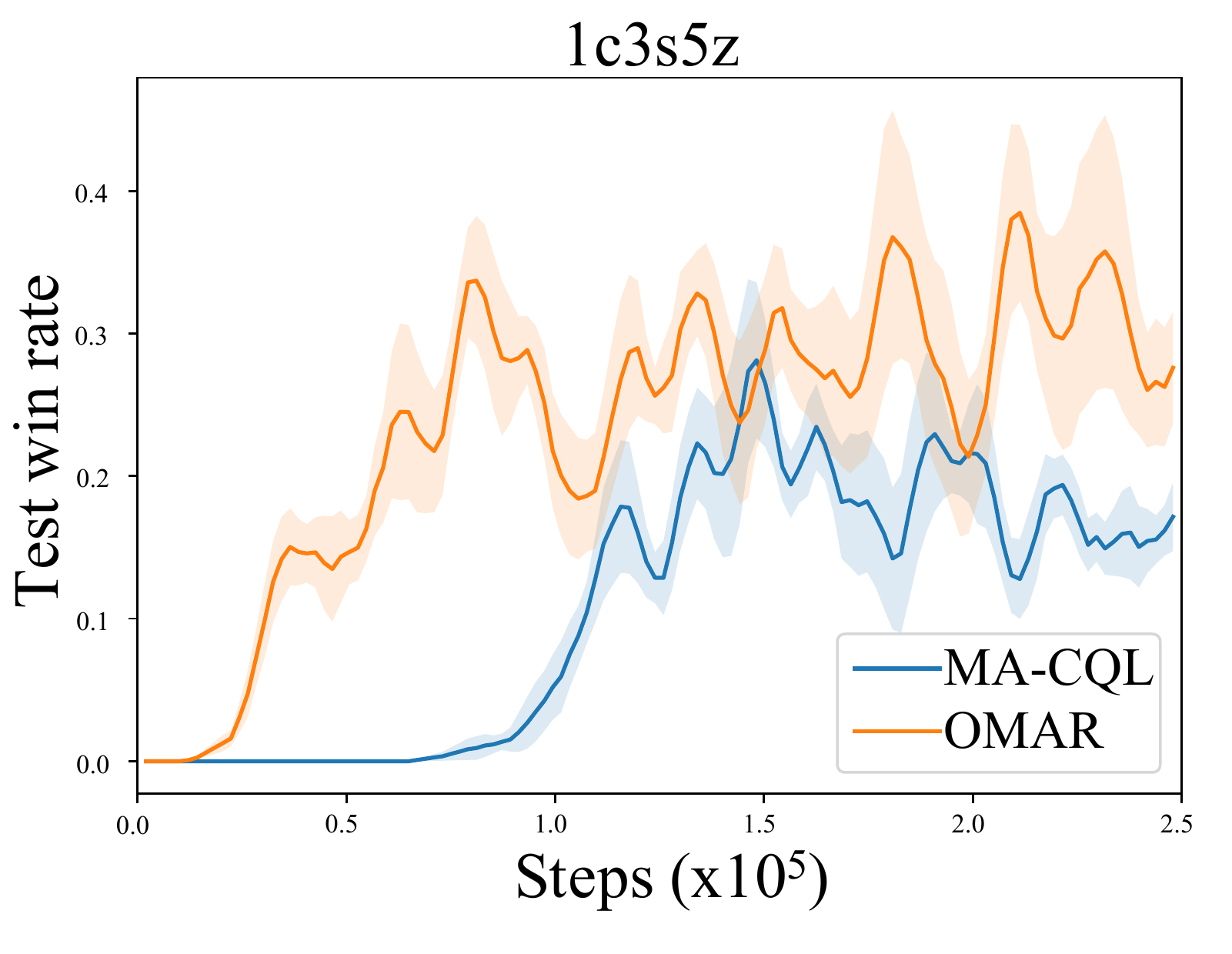}
\includegraphics[width=0.24\linewidth]{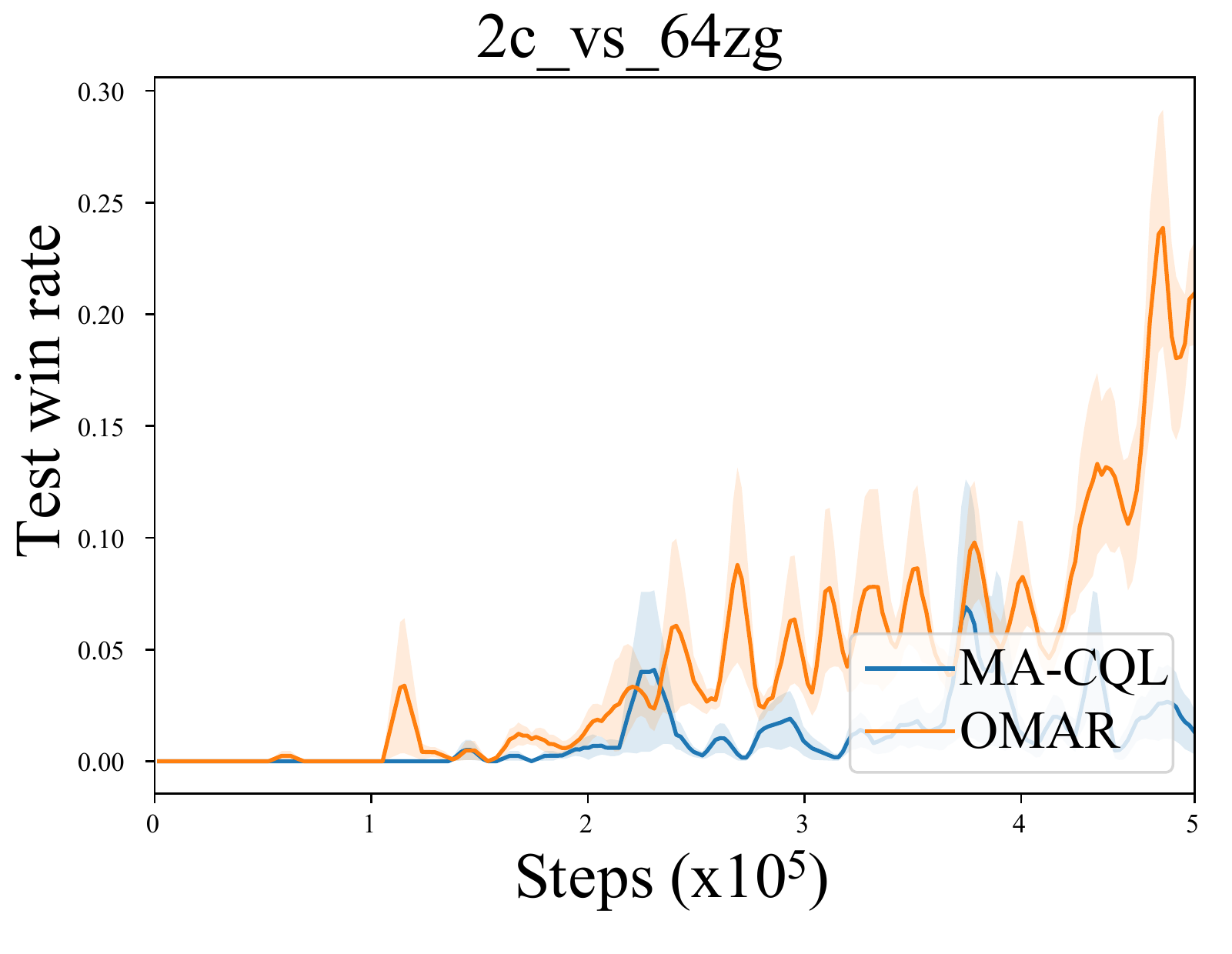}
\caption{Comparison of test win rates in StarCraft II micromanagement tasks.}
\label{fig:smac}
\end{figure}

\subsection{Compatibility in Single-Agent D4RL Environments}\label{app:d4rl}
Besides the single-agent setting of the Spread task we have studied in Figure \ref{fig:online_discuss}, we also evaluate the effectiveness of our method in single-agent tasks in the Maze2D domain from the D4RL benchmark~\citep{fu2020d4rl}. 
Table \ref{tab:d4rl} shows the comparison results of each method in an increasing order of complexity of the maze including umaze, medium, and large.
Based on the results in Table \ref{tab:d4rl} and Figure \ref{fig:online_discuss}, we find that OMAR also outperforms CQL, indicating that OMAR is also compatible for single-agent control as discussed in the previous section.

\begin{table}[!h]
\caption{Averaged normalized score of OMAR and baselines in the single-agent Maze2D domain.}
\centering
\begin{tabular}{cccc}
\toprule
& umaze & medium & large \\
\midrule
TD3+BC & $41.1\pm4.9$ & $75.5\pm27.1$ & $103.9\pm31.4$ \\
ICQ & $4.8\pm3.8$ & $13.0\pm7.9$ & $9.2\pm20.0$ \\
CQL & $109.8\pm23.9$ & $106.4\pm11.0$ & $94.6\pm44.6$ \\
OMAR & $\textbf{124.7}\pm7.6$ & $\textbf{125.7}\pm12.3$ & $\textbf{157.7}\pm12.3$ \\
\bottomrule
\end{tabular}
\label{tab:d4rl}
\end{table} 
\section{Related Work}
\paragraph{Offline reinforcement learning.} 
Many recent papers achieve improvements in offline RL~\citep{wu2019behavior,kumar2019stabilizing,yu2020mopo,kidambi2020morel,kostrikov2021offline,kostrikov2021offlineb} that address the extrapolation error.
Behavior regularization typically compels the learning policy to stay close to the behavior policy.
Yet, its performance relies heavily on the quality of the dataset. 
Critic regularization approaches typically add a regularizer to the critic loss which pushes down Q-values for actions sampled from a given policy~\citep{kumar2020conservative}. As discussed above, it can be difficult for the actor to best leverage the global information in the critic as policy gradient methods are prone to local optima, which is important in the offline multi-agent setting.

\paragraph{Multi-agent reinforcement learning.}
A number of multi-agent policy gradient algorithms train agents based on centralized value functions~\citep{lowe2017multi,foerster2018counterfactual,rashid2018qmix,yu2021surprising} while another line of research focuses on decentralized training~\citep{de2020independent}.
\citet{yang2021believe} show that the extrapolation error in offline RL can be more severe in the multi-agent setting than the single-agent case due to the exponentially sized joint action space w.r.t. the number of agents. In addition, it presents a critical challenge in the decentralized setting when the datasets for each agent only consist of its own action instead of the joint action~\citep{jiang2021offline}. \citet{jiang2021offline} address the challenges based on the behavior regularization BCQ~\citep{fujimoto2019off} algorithm while \citet{yang2021believe} propose to estimate the target value based on the next action from the dataset, where both methods can largely depend on the quality of the dataset. OMAR do not restricts its value estimate based only on seen state-action pairs in the dataset, and performs well in datasets with differnt quality.

\paragraph{Zeroth-order optimization method.} It has been recently shown in~\citep{such2017deep,conti2017improving,mania2018simple} that evolutionary strategies (ES) emerge as another paradigm for continuous control. Recent research shows that it has the potential to combine RL with ES to reap the best of both worlds~\citep{khadka2018evolution,pourchot2018cem} in the high-dimensional parameter space for the actor.
\citet{sun2020zeroth} replace the policy gradient update via supervised learning based on sampled noises from random shooting.
\citet{kalashnikov2018qt,lim2018actor,simmons2019q,peng2020facmac} extend Q-learning based approaches to handle continuous action space based on the popular cross-entropy method (CEM) in ES. 
\section{Conclusion}
In this paper, we study the important and challenging offline multi-agent RL setting, where we identify that directly extending current conservatism-based RL algorithms to offline multi-agent scenarios results in severe performance degradation along with an increasing number of agents through empirical analysis. We propose a simple yet effective method, OMAR, to tackle the problem by combining the first-order policy gradient with the zeroth-order optimization methods. We find that OMAR successfully helps the actor escape from bad local optima and consequently find better actions. Extensive experiments show that OMAR significantly outperforms state-of-the-art baselines on a variety of multi-agent control tasks. Interesting future directions include theoretical study for our identified problem in the offline multi-agent case with deep neural networks
, utilizing a more general class of distributions for the sampling mechanism, and the application of OMAR to other multi-agent RL methods.
\appendix
\section{Additional Results for Spread}
\subsection{Results with Larger Learning Rates and Number of Updates of Actors in MA-CQL} \label{sec:detailed_mot}
Table \ref{tab:increase_lr} shows the result of MA-CQL with larger learning rates, where we also include results for using smaller learning rates for reference. Table \ref{tab:increase_updates} demonstrates the result of MA-CQL with larger numbers of updates for actors.
\begin{table}[!h]
\centering
\caption{Performance of MA-CQL with larger learning rate for the actor.}
\label{tab:increase_lr}
\begin{tabular}{ccccccc}
\toprule
Learning rate & $5e-4$ & $1e-3$ & $5e-3$ & $1e-2$ & $5e-2$ & $1e-1$ \\
\midrule
Performance & $152.3\pm17.1$ & $164.0\pm14.5$ & $256.2\pm34.2$ & $267.9\pm19.0$ & $202.0\pm38.9$ & $100.1\pm36.4$  \\
\bottomrule
\end{tabular}
\end{table}

\begin{table}[!h]
\centering
\caption{Performance of MA-CQL with larger number of updates for the actor.}
\label{tab:increase_updates}
\begin{tabular}{cccc}
\toprule
\# Updates & $1$ & $5$ & $20$ \\
\midrule
Performance & $267.9\pm19.0$ & $278.6\pm14.8$ & $263.7\pm23.1$\\
\bottomrule
\end{tabular}
\end{table}

\subsection{The Performance of MA-CQL in a Non-Cooperative Version of the Multi-Agent Spread Task}\label{app:cooperation}
We consider a non-cooperative version of the Spread task in Figure \ref{fig:spread_env} which involves $n$ agents and $n$ landmarks, where each of the agents aims to navigate to its own unique target landmark. In contrast to the Spread task that requires cooperation, the reward function for each agent only depends on its distance to its target landmark. This is a variant of Spread that consists of multiple independent learning agents, and the performance is measured by the average return over all agents.

\begin{figure}[!h]
\centering
\includegraphics[width=0.25\linewidth]{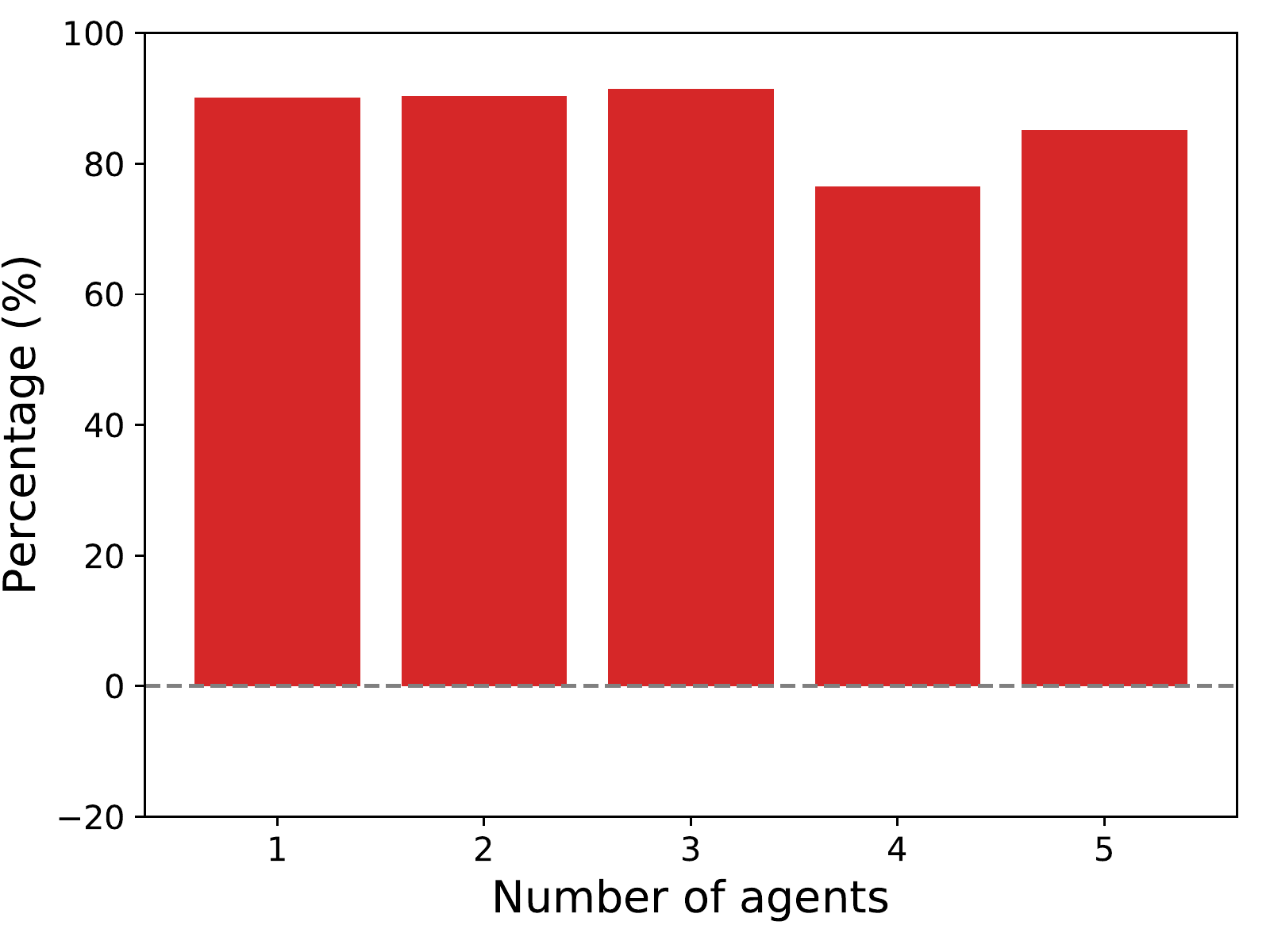}
\caption{Performance improvement percentage of MA-CQL over the behavior policy with a varying number of agents in a non-cooperative version of the Spread task.}
\label{fig:noncoop}
\end{figure}

Figure \ref{fig:noncoop} shows the result of the performance improvement percentage of MA-CQL over the behavior policy in the independent Spread task.
As shown, the performance of CQL does not degrade with an increasing number of agents in this setting that does not require cooperation, unlike a dramatic performance decrease in the cooperative Spread task in Figure \ref{fig:simple_navi}(b). The result further confirms that the issue we discovered is due to the failure of coordination.

\section{Proof of Theorem 3.1}\label{app:proof}
\textbf{Theorem 3.1.}
\emph{
Let $\pi_i^*$ denote the policy obtained by optimizing Eq. (\ref{eq:rac}), $D(\pi_i,\hat{\pi}_{\beta_i})(o_i) = \frac{1-\hat{\pi}_{\beta_i}(\pi_i(o_i)|o_i)}{\hat{\pi}_{\beta_i}(\pi_i(o_i)|o_i)}$, and $d^{\pi_i}(o_i)$ denote the marginal discounted distribution of observations of policy $\pi_i$. Then, we have that $J(\pi_i^*) - J(\hat{\pi}_{\beta_i}) \geq  \frac{\alpha}{1-\gamma} \mathbb{E}_{o_i\sim d^{\pi_i^*}(o_i)}\left[D(\pi_i^*,\hat{\pi}_{\beta_i})(o_i)\right]+ \frac{\tau}{1-\tau}\mathbb{E}_{o_i\sim d^{\pi_i^*}(o_i)} \left[ (\pi_i^*(o_i)-\hat{a}_i)^2 \right]- \frac{\tau}{1-\tau}\mathbb{E}_{o_i\sim d^{\hat{\pi}_{\beta_i}}(o_i),a_i\sim\hat{\pi}_{\beta_i}} \left[ (a_i-\hat{a}_i)^2 \right].$
}

\begin{proof}
For OMAR, we have the following iterative update for agent $i$:
\begin{equation}
\begin{split}
\hat{Q}_i^{k+1} \leftarrow {\arg\min}_{Q_i} & \alpha \mathbb{E}_{{o_i} \sim \mathcal{D}_i} \left[ \mathbb{E}_{{a_i}\sim {\tilde{\pi}}_{i}({a_i} \mid {o_i})} \left[Q_i({o_i}, {a_i})\right] - \mathbb{E}_{{a_i} \sim \hat{\pi}_{\beta_i}({a_i} \mid {o_i})} \left[Q_i({o_i}, {a_i})\right] \right] \\
& + \frac{1}{2} \mathbb{E}_{{o_i}, {a_i}, {o_i}^{\prime} \sim \mathcal{D}}\left[\left(Q_i(o_i, a_i)-\hat{\mathcal{B}}^{\pi_{i}} \hat{Q}_{i}^{k}(o_i, a_i)\right)^{2}\right],
\end{split}
\label{cqlnew}
\end{equation}
where $\tilde{\pi}_{i}({a_i}|{o_i})=1$ if and only if $a_i=\pi_i(o_i).$

Let $\hat{Q}_i^{k+1}$ be the fixed point of solving Equation (\ref{cqlnew}) by setting the derivative of Eq. (\ref{cqlnew}) with respect to $Q_i$ to be $0$, then we have that
\begin{equation}
\hat{Q}_i^{k+1}(o_i,a_i) = \hat{\mathcal{B}}^{\pi_{i}}\hat{Q}_{i}^{k}(o_i,a_i)-\alpha \left(\frac{I_{a_i=\pi_i(o_i)}}{\hat{\pi}_{\beta_i}(a_i|o_i)}-1 \right),
\end{equation}
where $I$ is the indicator function.

Denote $D(\pi_i,\hat{\pi}_{\beta_i})(o_i) = \frac{1}{\hat{\pi}_{\beta_i}(\pi_i(o_i)|o_i)}-1$, and we obtain the difference between the value function $\hat{V}_i(o_i)$ and the original value function as:
\begin{equation}
\hat{V}_i(o_i)={V}_i(o_i)-\alpha D(\pi_i,\hat{\pi}_{\beta_i})(o_i),
\end{equation}

Then, the policy that minimizes the loss function defined in Eq. (\ref{eq:rac}) is equivalently obtained by maximizing
\begin{equation}
\begin{split}
(1-\tau) \left(J(\pi_i)-\alpha \frac{1}{1-\gamma} \mathbb{E}_{o_i\sim d^{\pi_i}_{\hat{M}_i}(o_i)} \left[D(\pi_i,\hat{\pi}_{\beta_i})(o_i)\right] \right) - \tau \mathbb{E}_{o_i\sim d^{\pi_i}_{\hat{M}_i}(o_i)} \left[ (\pi_i(o_i)-\hat{a}_i)^2 \right].
\end{split}
\end{equation}

Therefore, we obtain that
\begin{equation}
\begin{split}
&(1-\tau) \left(J(\pi_i^*)-\alpha \frac{1}{1-\gamma} \mathbb{E}_{o_i\sim d^{\pi_i^*}_{\hat{M}_i}(o_i)} \left[D(\pi_i^*,\hat{\pi}_{\beta_i})(o_i) \right] \right)
-\tau \mathbb{E}_{o_i\sim d^{\pi_i^*}_{\hat{M}_i}(o_i)} \left[(\pi_i^*(o_i)-\hat{a}_i)^2 \right]\\
\geq & (1-\tau) J(\hat{\pi}_{\beta_i})
-\tau \mathbb{E}_{o_i\sim d^{\hat{\pi}_{\beta_i}}_{\hat{M}_i}(o_i),{a_i} \sim \hat{\pi}_{\beta_i}({a_i} \mid {o_i})} \left[ (a_i-\hat{a}_i)^2 \right].
\end{split}
\label{eq:thm1_final}
\end{equation}
Then, from Eq. (\ref{eq:thm1_final}) we obtain the result.
\end{proof}

\section{Experimental Details}

\subsection{Experimental Setup}\label{app:exp}
\subsubsection{Tasks.}\label{sec:task}
\begin{figure}[!h]
\centering
\subfloat[Cooperative navigation.]{\includegraphics[width=0.25\linewidth]{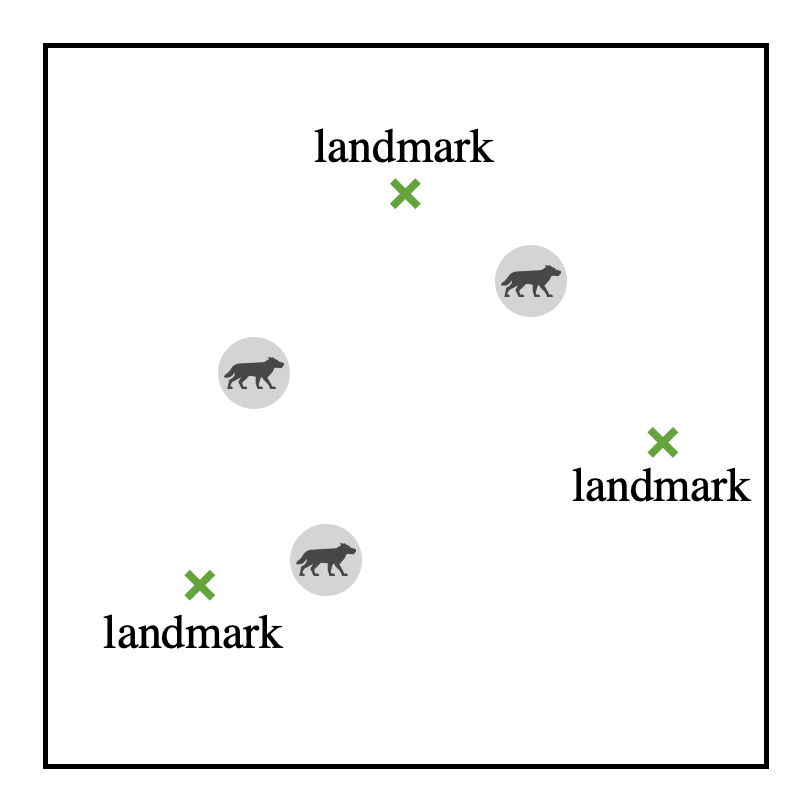}}
\subfloat[Predator-prey.]{\includegraphics[width=0.25\linewidth]{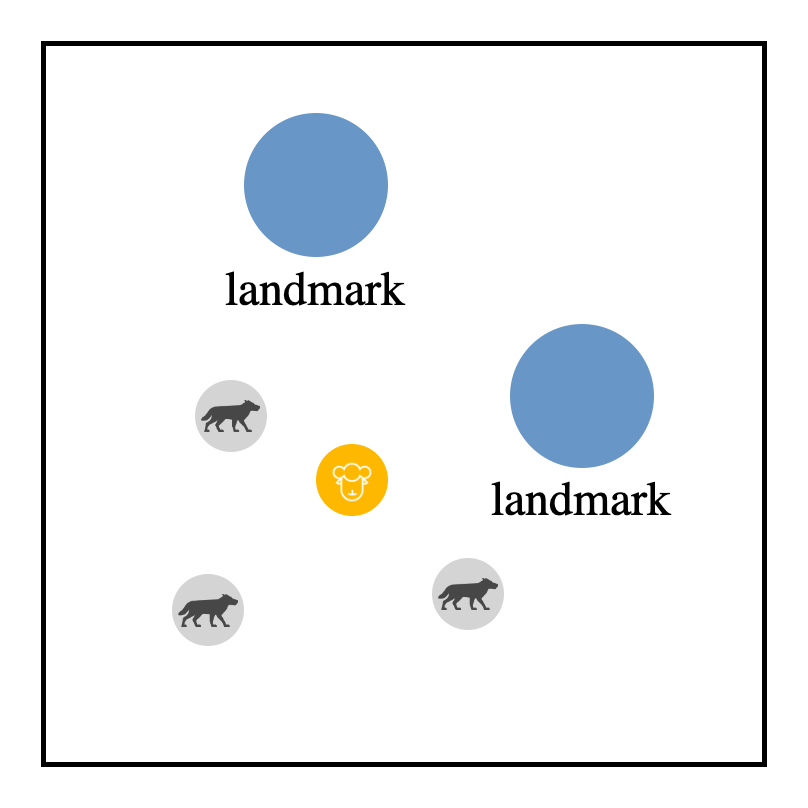}}
\subfloat[World.]{\includegraphics[width=0.25\linewidth]{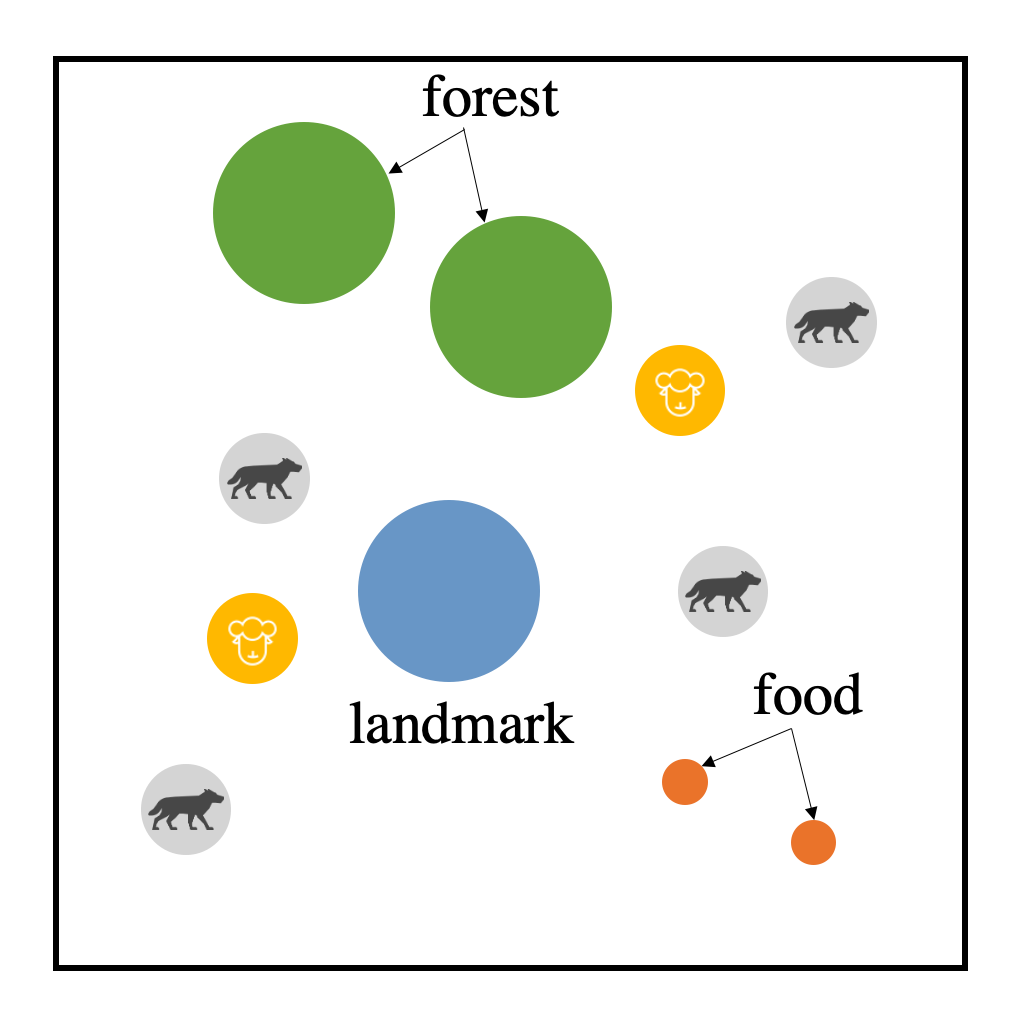}}
\subfloat[Two-agent HalfCheetah.]{\includegraphics[width=0.25\linewidth]{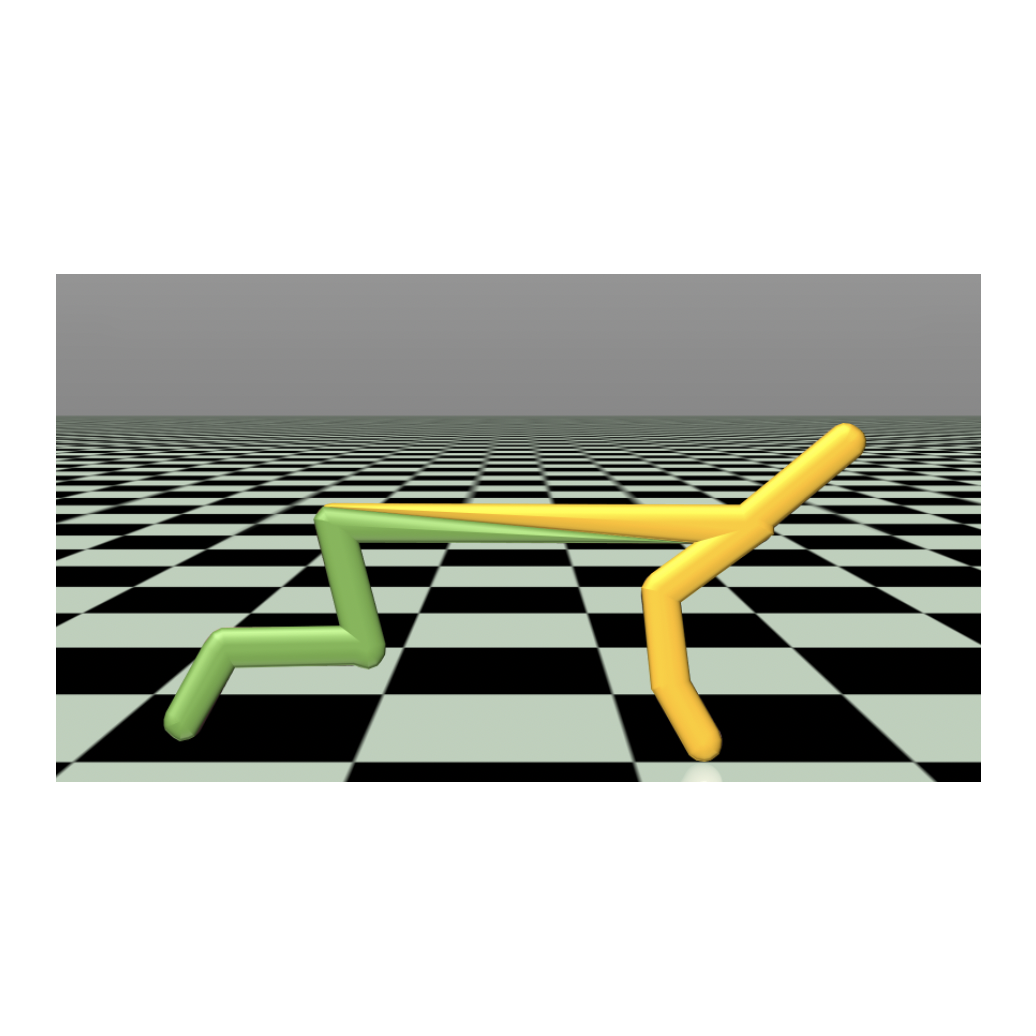}}\\
\subfloat[Maze2D-Umaze-Dense.]{\includegraphics[width=0.25\linewidth]{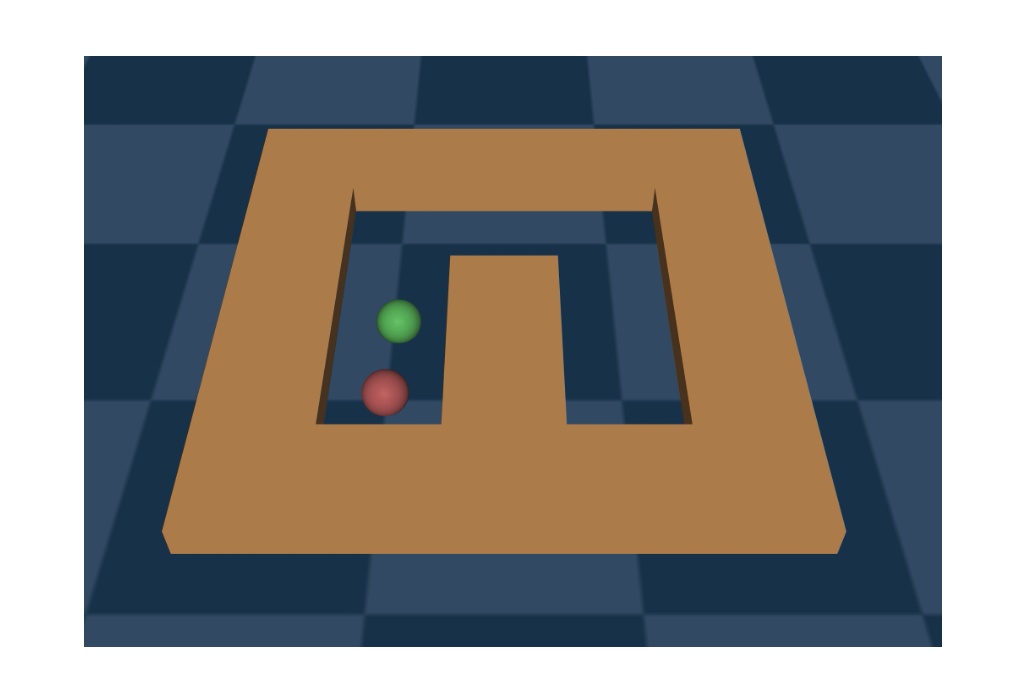}}
\subfloat[Maze2D-Medium-Dense.]{\includegraphics[width=0.25\linewidth]{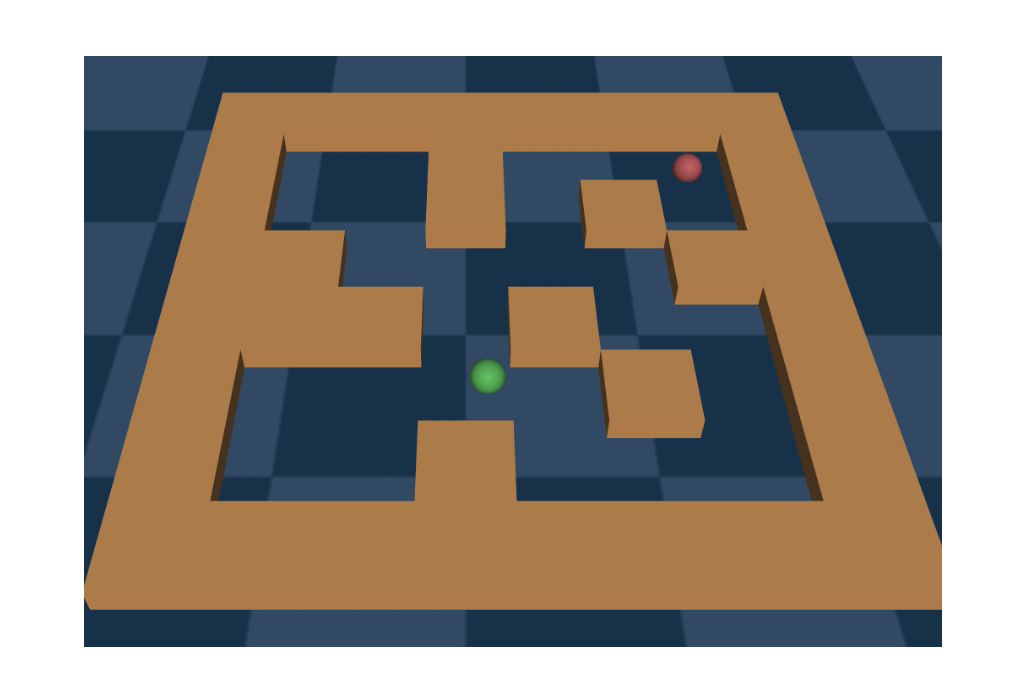}}
\subfloat[Maze2D-Large-Dense.]{\includegraphics[width=0.25\linewidth]{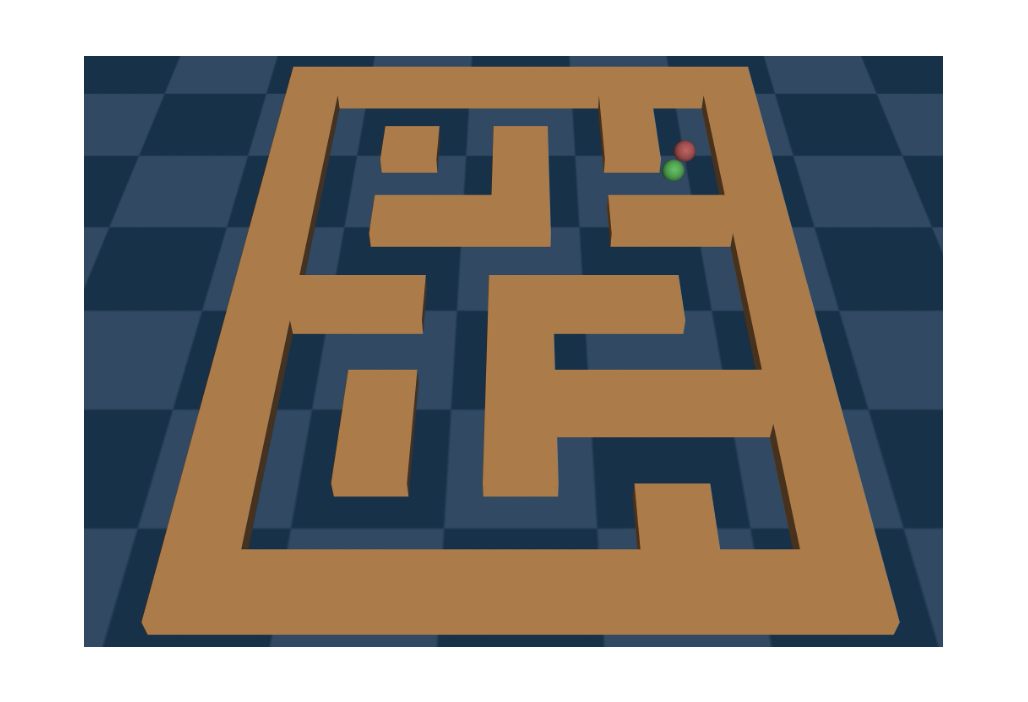}}
\caption{Multi-agent particle environments and Multi-Agent HalfCheetah.}
\label{fig:mpe_env}
\end{figure}

We adopt the open-source implementations for multi-agent particle environments\footnote{\url{https://github.com/openai/multiagent-particle-envs}} from~\citep{lowe2017multi}, Multi-Agent MuJoCo\footnote{\url{https://github.com/schroederdewitt/multiagent_mujoco}} from~\citep{peng2020facmac}, and StarCraft II Micromanagement Benchmark\footnote{\url{https://github.com/oxwhirl/smac}} from~\citep{samvelyan2019starcraft}.
Figures \ref{fig:mpe_env}(a)-(c) illustrate tasks from multi-agent particle environments.
The cooperative navigation task includes $3$ agents and $3$ landmarks, where agents are rewarded based on the distance to the landmarks and penalized for colliding with each other. Thus, it is important for agents to cooperate to cover all landmarks without collision. In predator-prey, $3$ predators aim to catch the prey. The predators need to cooperate to surround and catch the prey as the predators are slower than the prey. The world task involves $4$ slower cooperating agents that aim to catch $2$ faster adversaries, where adversaries desire to eat foods while avoiding being captured.
The two-agent HalfCheetah task is shown in Figure \ref{fig:mpe_env}(d) while Figures \ref{fig:mpe_env}(e)-(g) illustrate the Maze2D environments from the D4RL\footnote{\url{https://github.com/rail-berkeley/d4rl}} benchmark~\citep{fu2020d4rl}.
The expert and random scores for cooperative navigation, predator-prey, world, and two-agent HalfCheetah are $\{516.8, 159.8\}$, $\{185.6, -4.1\}$, $\{79.5, -6.8\}$, and $\{3568.8, -284.0\}$, respectively.
Tested maps in StarCraft II micromanagement benchmark are summarized in Table \ref{table:smac_env}.

\begin{table}[!h]
  \centering
  \caption{Specs of tested maps in the StarCraft II micromanagement benchmark.}
  \begin{tabular}{lccc}
    \toprule
    {\bf Name} & {\bf Agents} & {\bf Enemies}\\
    \midrule
    2s3z & 2 Stalkers and 3 Zealots & 2 Stalkers and 3 Zealots \\
    3s5z & 3 Stalkers and 5 Zealots & 3 Stalkers and 5 Zealots \\
    1c3s5z & 1 Colossi, 3 Stalkers and 5 Zealots & 1 Colossi, 3 Stalkers and 5 Zealots \\
    2c\_vs\_64zg & 2 Colossi & 64 Zerglings \\
    \bottomrule
  \end{tabular}
  \label{table:smac_env}
\end{table}

\subsubsection{Baselines.}\label{sec:baselines}
All baseline methods are implemented based on an open-source implementation\footnote{\url{https://github.com/shariqiqbal2810/maddpg-pytorch}} from ~\citep{iqbal2019actor}, where we implement MA-TD3+BC\footnote{\url{https://github.com/sfujim/TD3_BC}}, MA-CQL\footnote{\url{https://github.com/aviralkumar2907/CQL}}, and MA-ICQ\footnote{\url{https://github.com/YiqinYang/ICQ}} based on open-source implementations with fine-tuned hyperparameters. For MA-CQL, we tune a best critic regularization coefficient from $\{0.1,0.5,1.0,5.0\}$ following~\citep{kumar2020conservative} for each task.
Specifically, we use the discount factor $\gamma$ of $0.99$. We sample a minibatch of $1024$ samples from the dataset for updating each agent's actor and critic using the Adam~\citep{kingma2014adam} optimizer with the learning rate to be $0.01$. The target networks for the actor and critic are soft updated with the update rate to be $0.01$. Both the actor and critic networks are feedforward networks consisting of two hidden layers with $64$ neurons per layer using ReLU activation.
For OMAR, the only hyperparameter that requires tuning is the regularization coefficient $\lambda$, where we use a smaller value for datasets with more diverse data distribution in random and medium-replay with a value of $0.5$, while we use a larger value for datasets with more narrow data distribution in medium and expert with values of $0.7$ and $0.9$ respectively.
As OMAR is insensitive to the hyperparameters of the sampling mechanism, we set them to a fixed set of values for all types of datasets in all tasks, where the number of iterations is $3$, the number of samples is $10$, the mean is $0.0$, and the standard deviation is $2.0$. For OMAR in the StarCraft II micromanagement benchmark, we follow the fine-tuned set of hyperparameters for MATD3 in ~\citep{peng2020facmac}. The code will be released upon publication of the paper.

\subsection{Learning Curves}\label{app:res}
\begin{figure}[!h]
\centering
\subfloat[CN-random]{\includegraphics[width=0.25\linewidth]{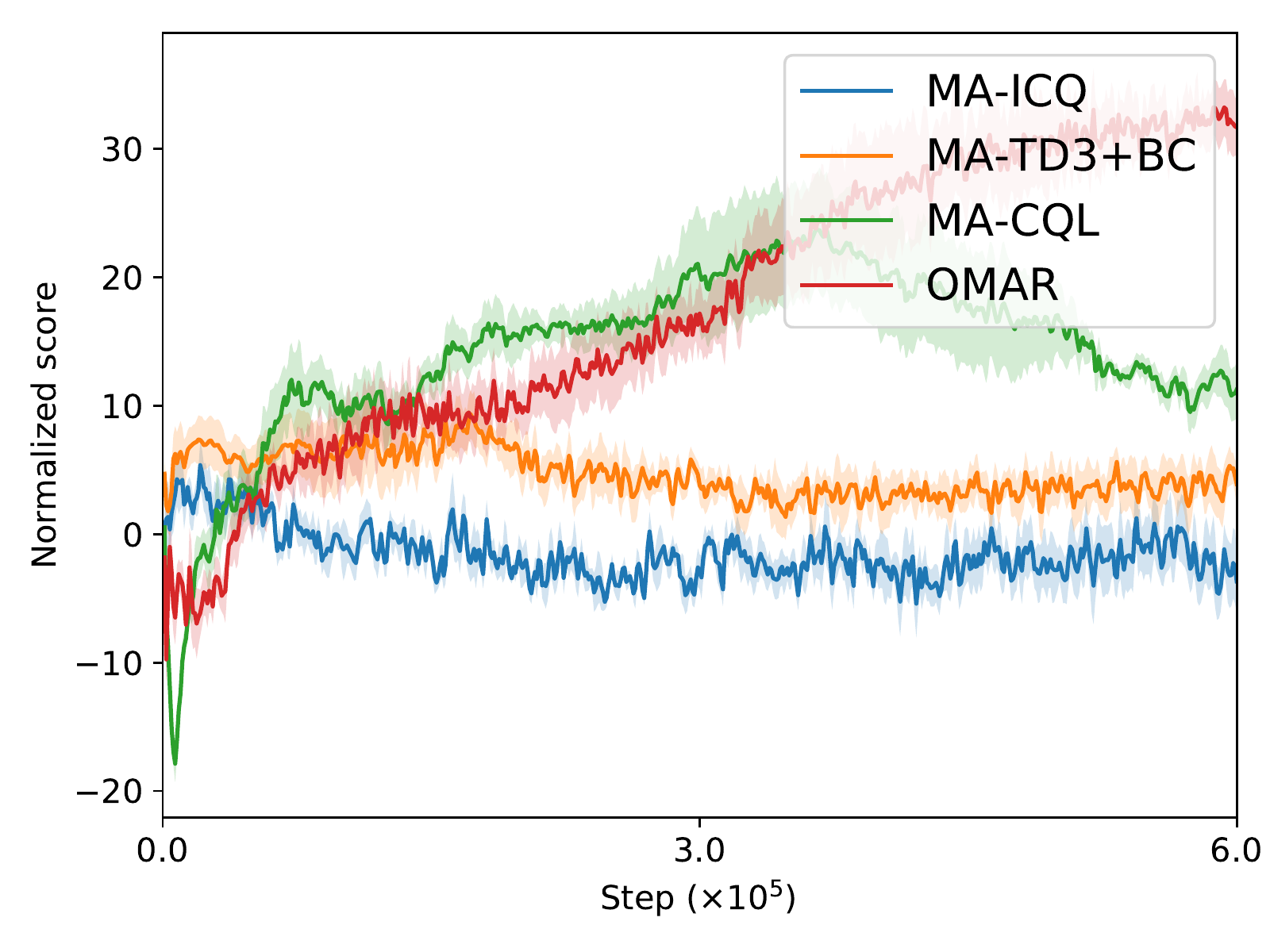}}
\subfloat[CN-medium-replay]{\includegraphics[width=0.25\linewidth]{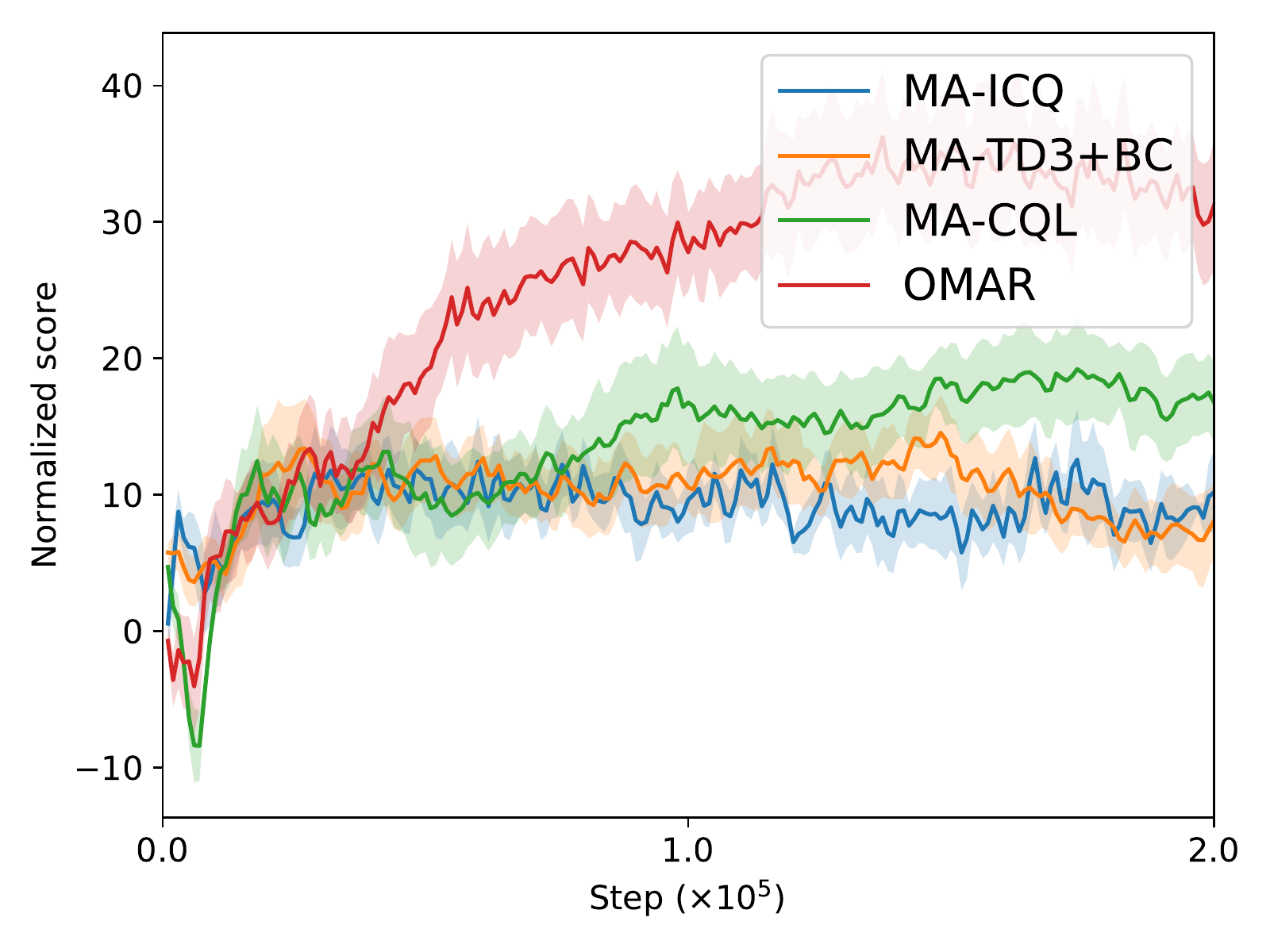}}
\subfloat[CN-medium]{\includegraphics[width=0.25\linewidth]{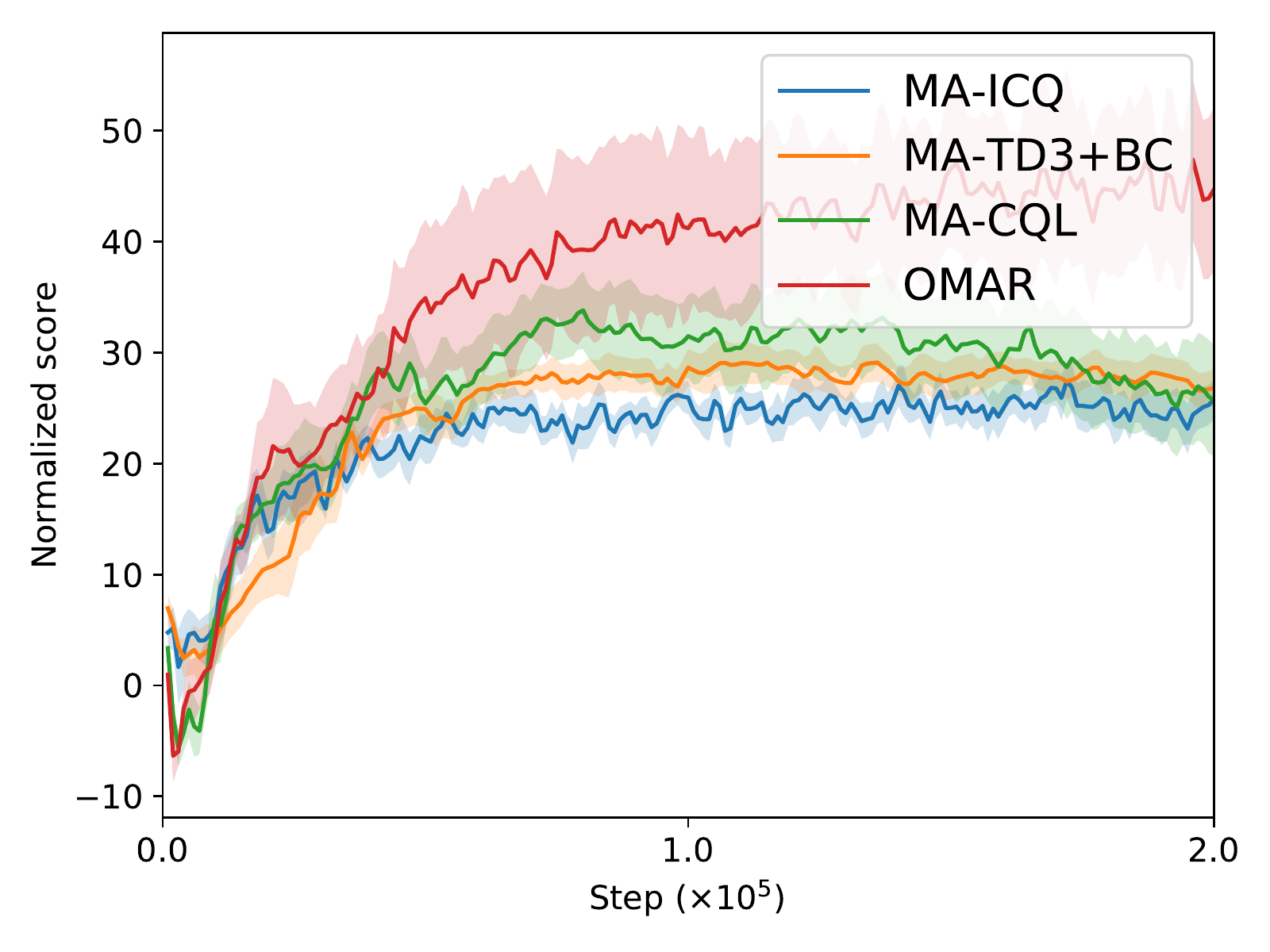}}
\subfloat[CN-expert]{\includegraphics[width=0.25\linewidth]{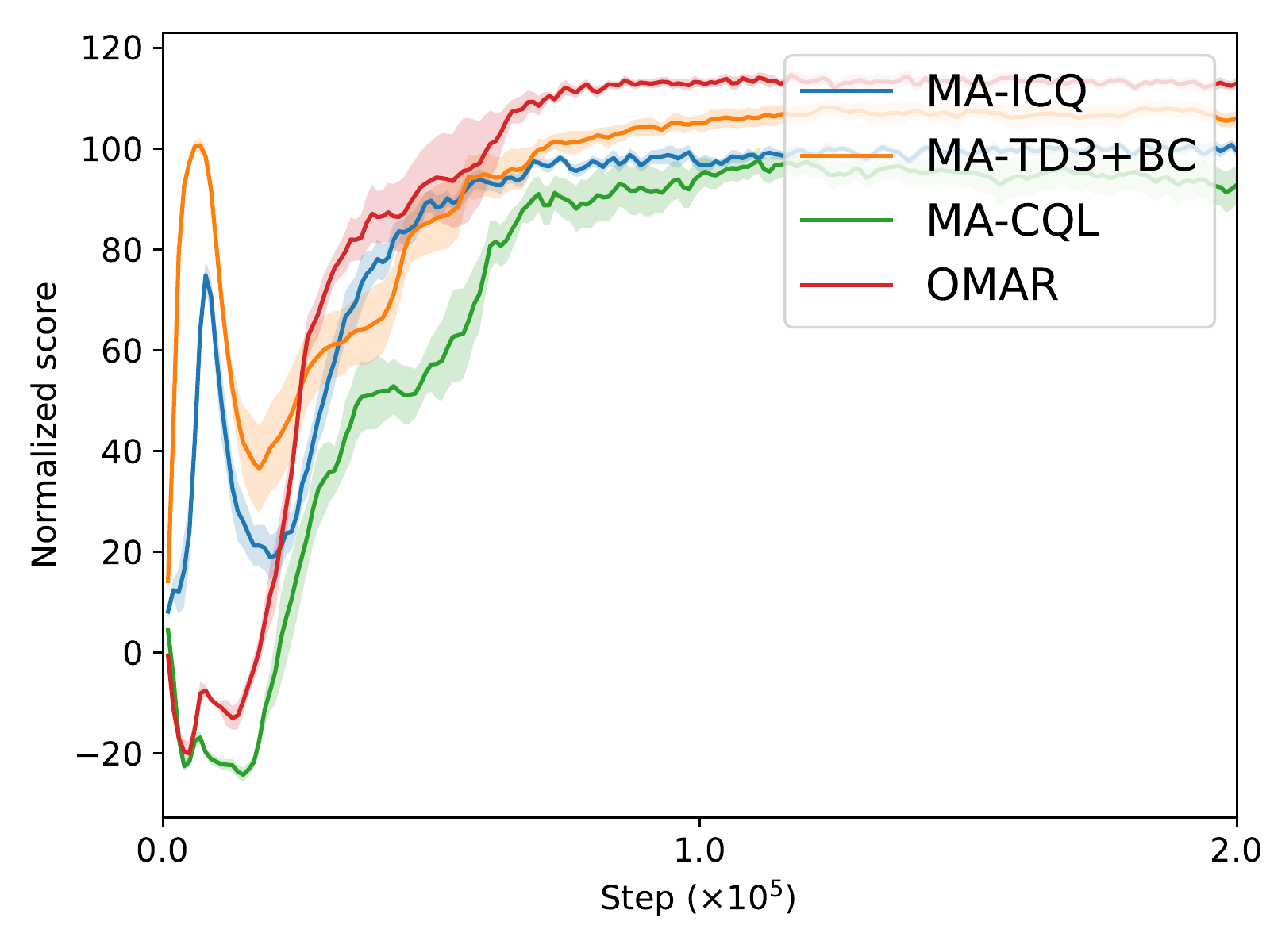}} \\
\subfloat[PP-random]{\includegraphics[width=0.25\linewidth]{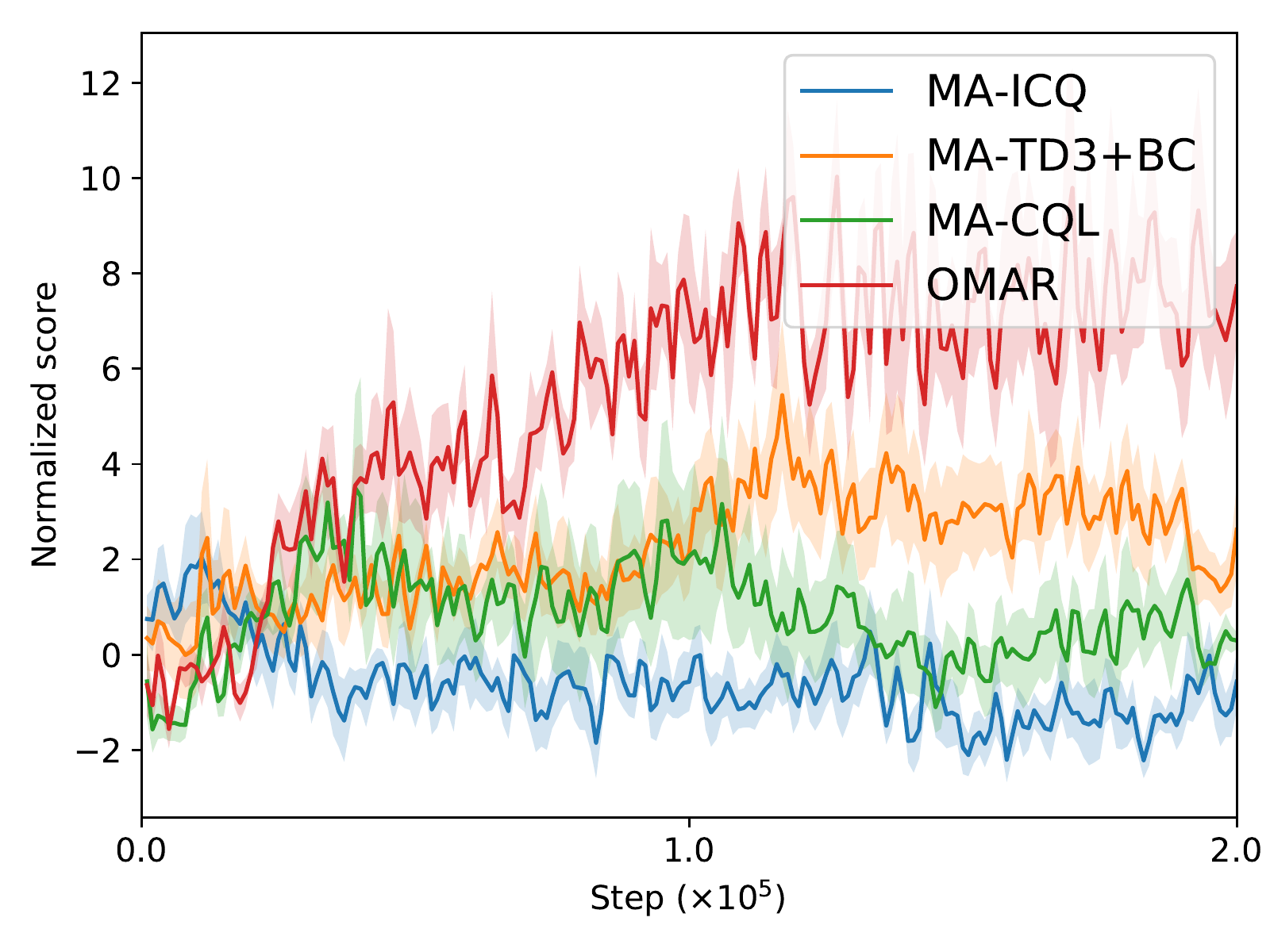}}
\subfloat[PP-medium-replay]{\includegraphics[width=0.25\linewidth]{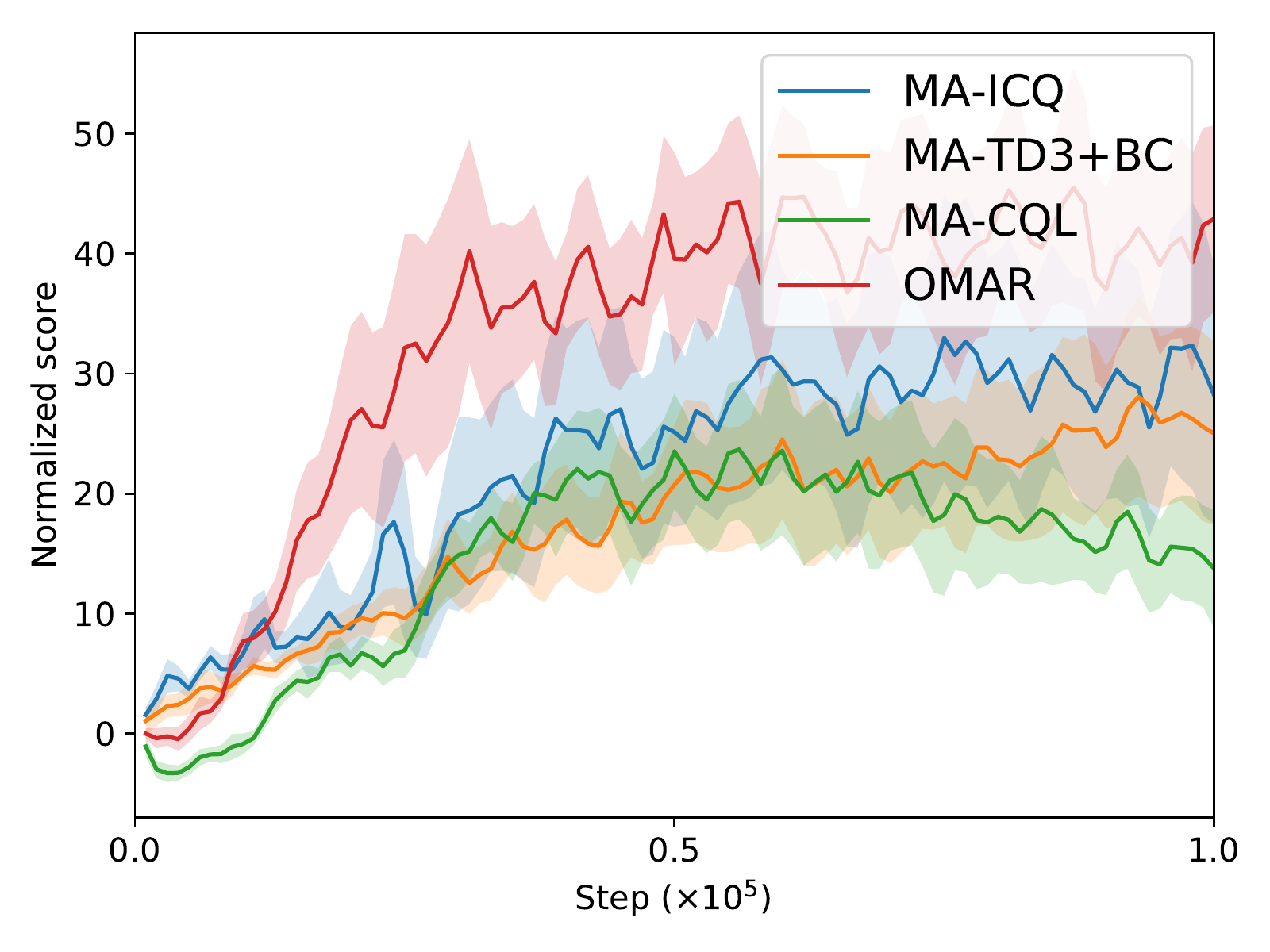}}
\subfloat[PP-medium]{\includegraphics[width=0.25\linewidth]{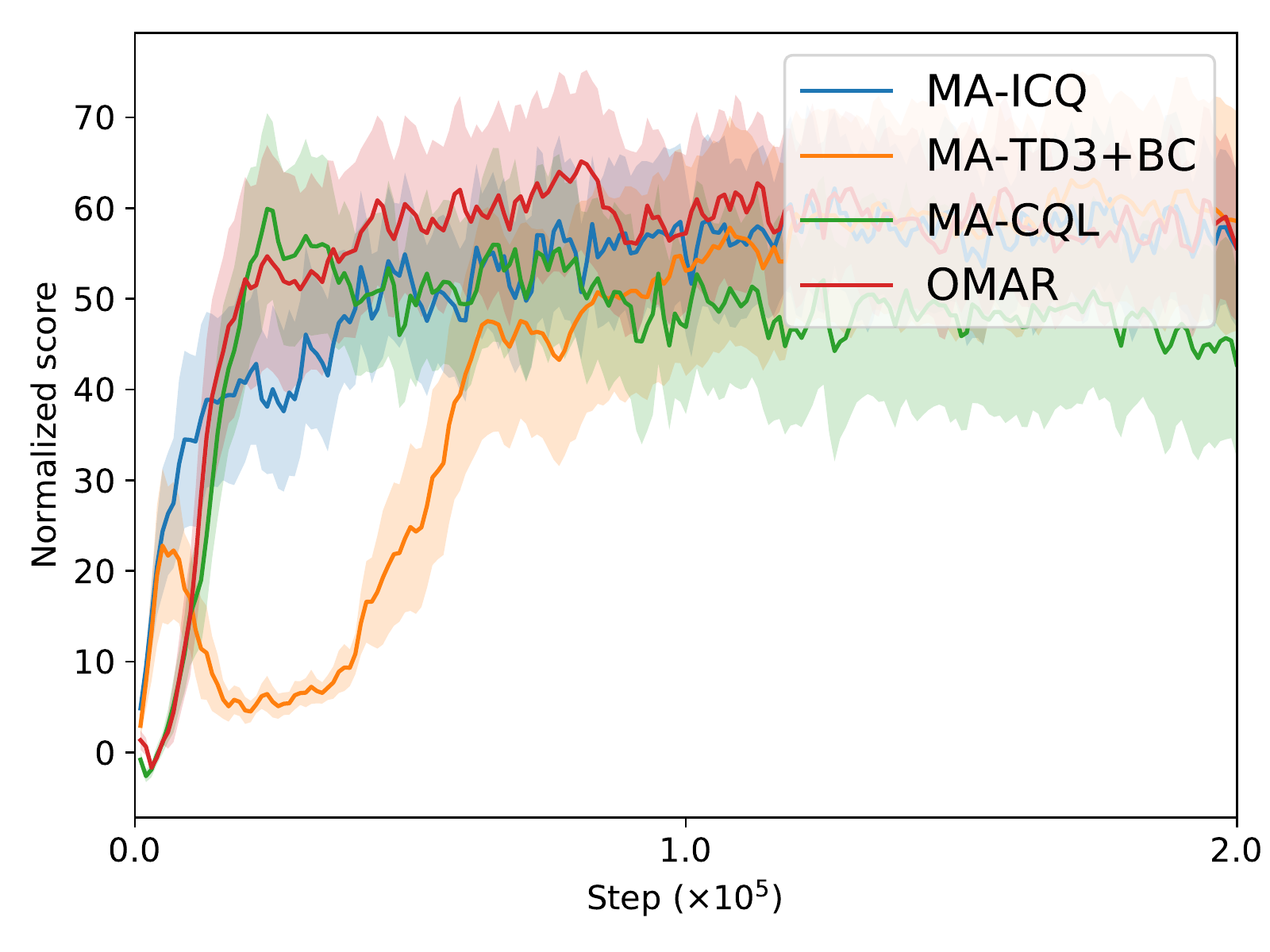}}
\subfloat[PP-expert]{\includegraphics[width=0.25\linewidth]{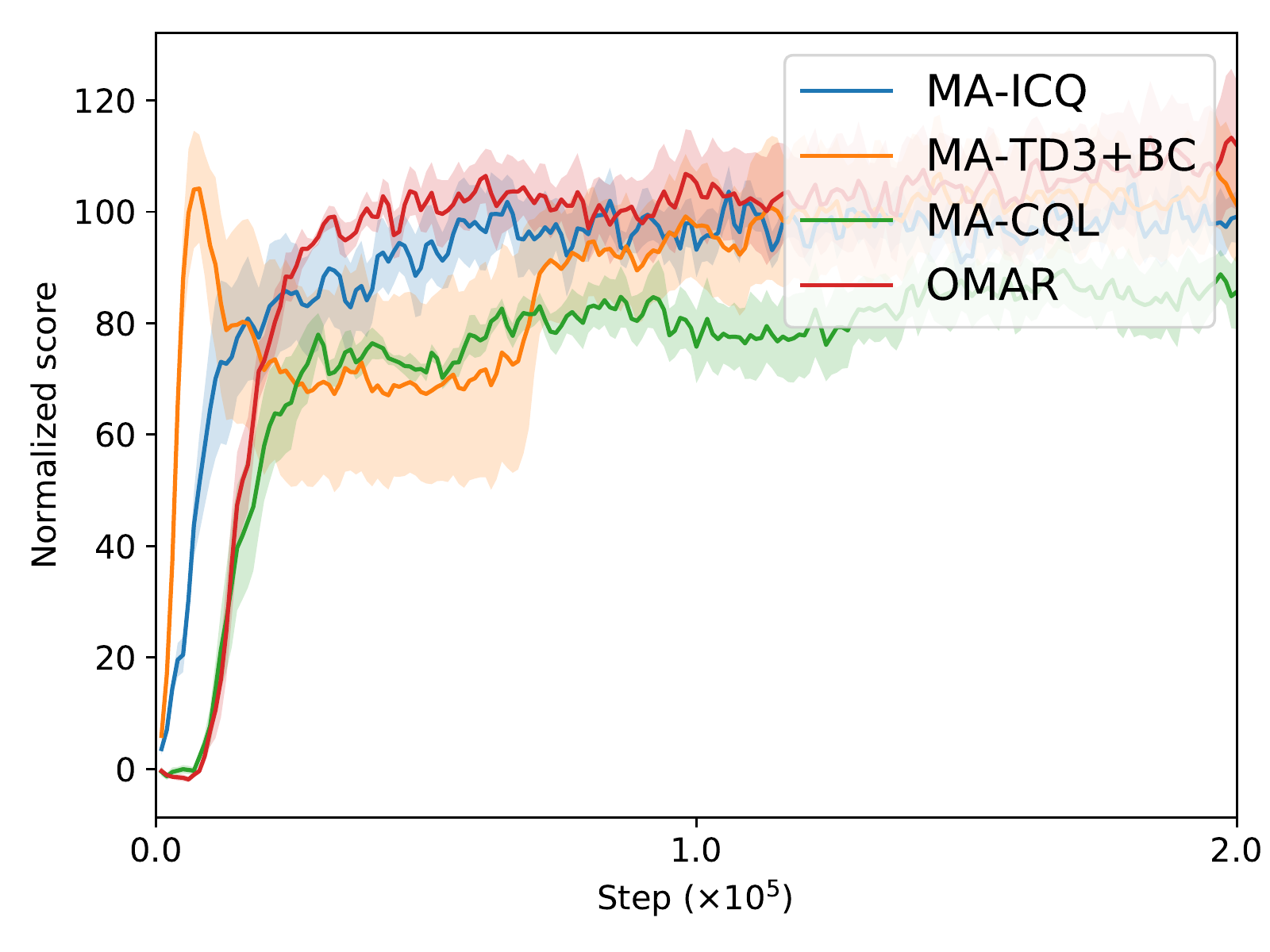}} \\
\subfloat[W-random]{\includegraphics[width=0.25\linewidth]{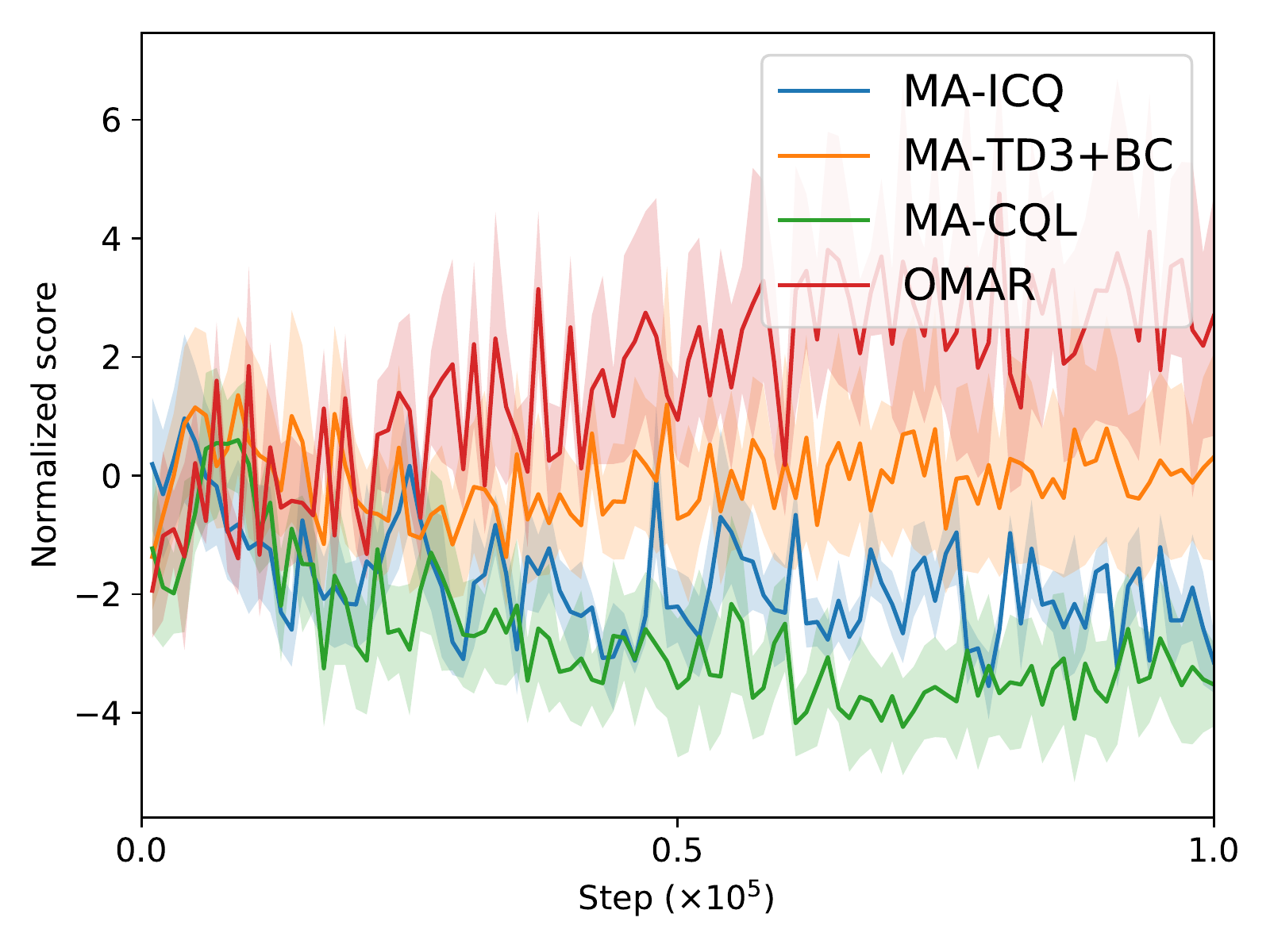}}
\subfloat[W-medium-replay]{\includegraphics[width=0.25\linewidth]{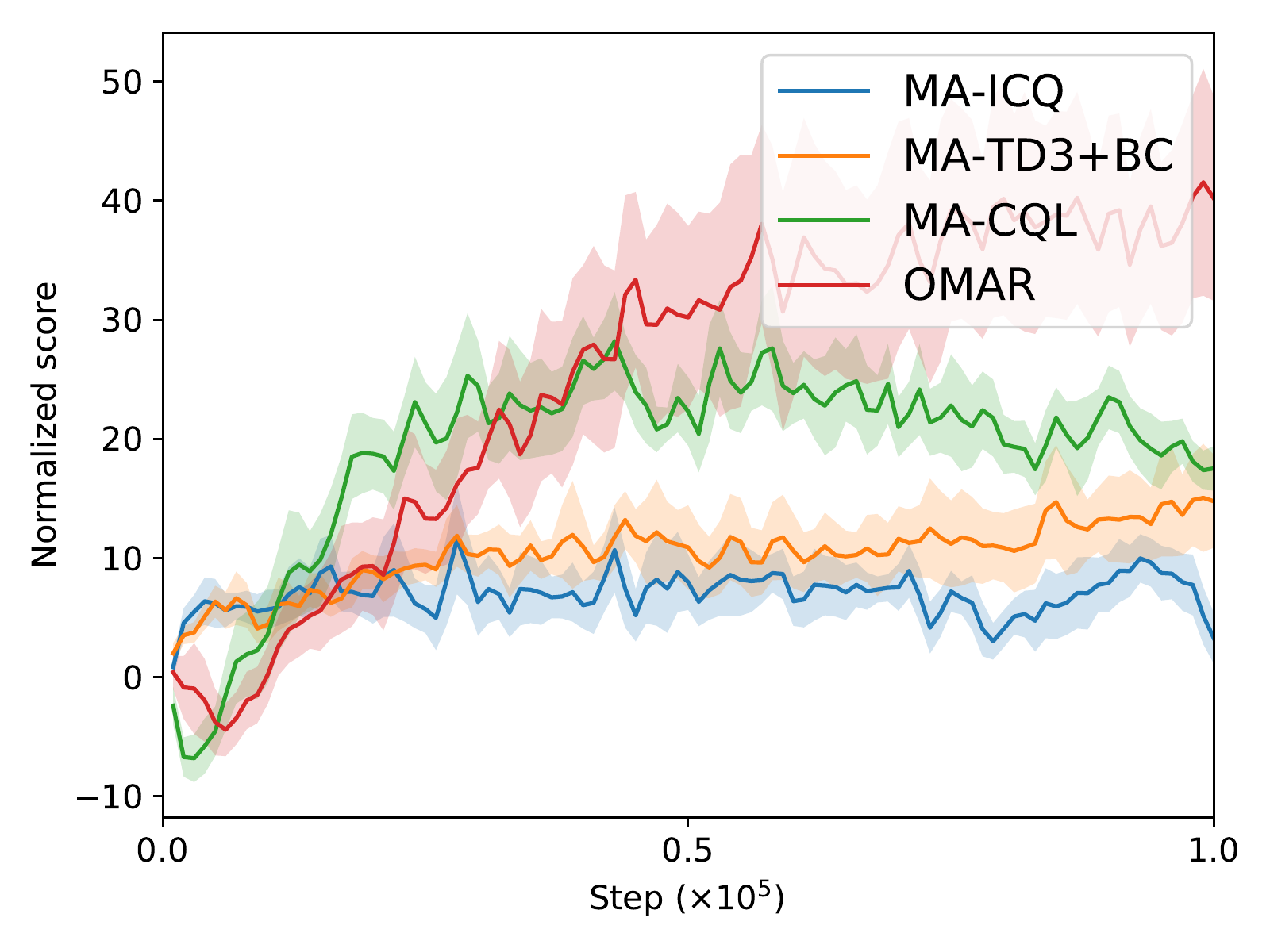}}
\subfloat[W-medium]{\includegraphics[width=0.25\linewidth]{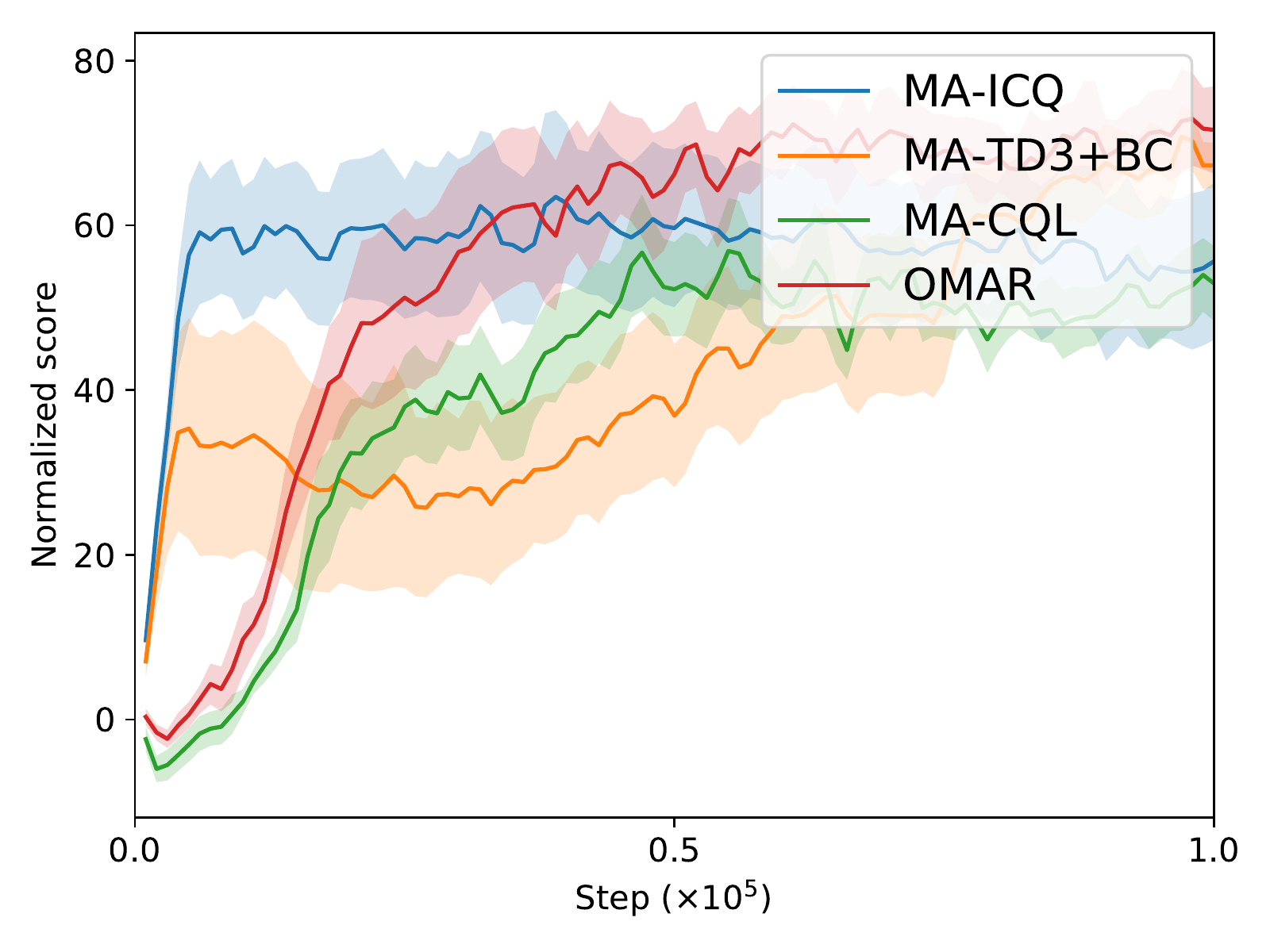}}
\subfloat[W-expert]{\includegraphics[width=0.25\linewidth]{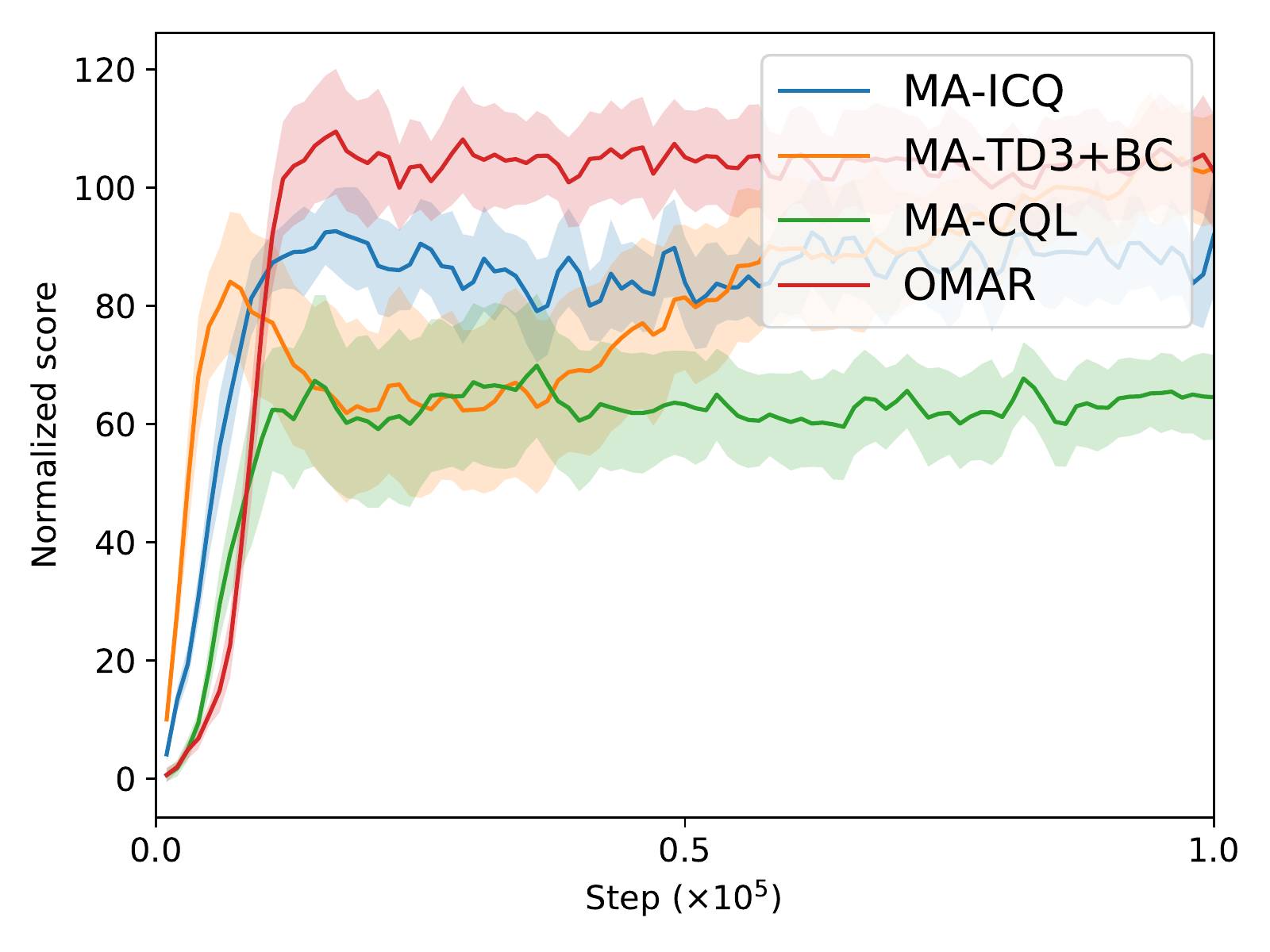}} \\
\caption{Learning curves of MA-ICQ, MA-TD3+BC, MA-CQL, and OMAR in multi-agent particle environments (CN, PP, and W is abbreviated for cooperative navigation, predator-prey, and world respectively).}
\label{fig:mpe_per}
\end{figure}

Figure \ref{fig:mpe_per} demonstrates the learning curves of MA-ICQ, MA-TD3+BC, MA-CQL and OMAR in different types of datasets in multi-agent particle environments, where the solid line and shaded region represent mean and standard deviation, respectively.

\subsection{Additional Ablation Study on the Effect of the Size of the Dataset} \label{app:size_ab}
In this section, we conduct an ablation study to investigate the effect of the size of the dataset following the experimental protocol in \citet{agarwal2020optimistic}.
We first generate a full replay dataset by recording all samples in the replay buffer encountered during the training course for $1$ million steps. Then, we randomly sample $N\%$ experiences from the full replay dataset and obtain several smaller datasets with the same data distribution, where $N\in\{0.1,1,10,20,50,100\}$. 

\begin{figure}[!h]
\centering
\includegraphics[width=0.3\linewidth]{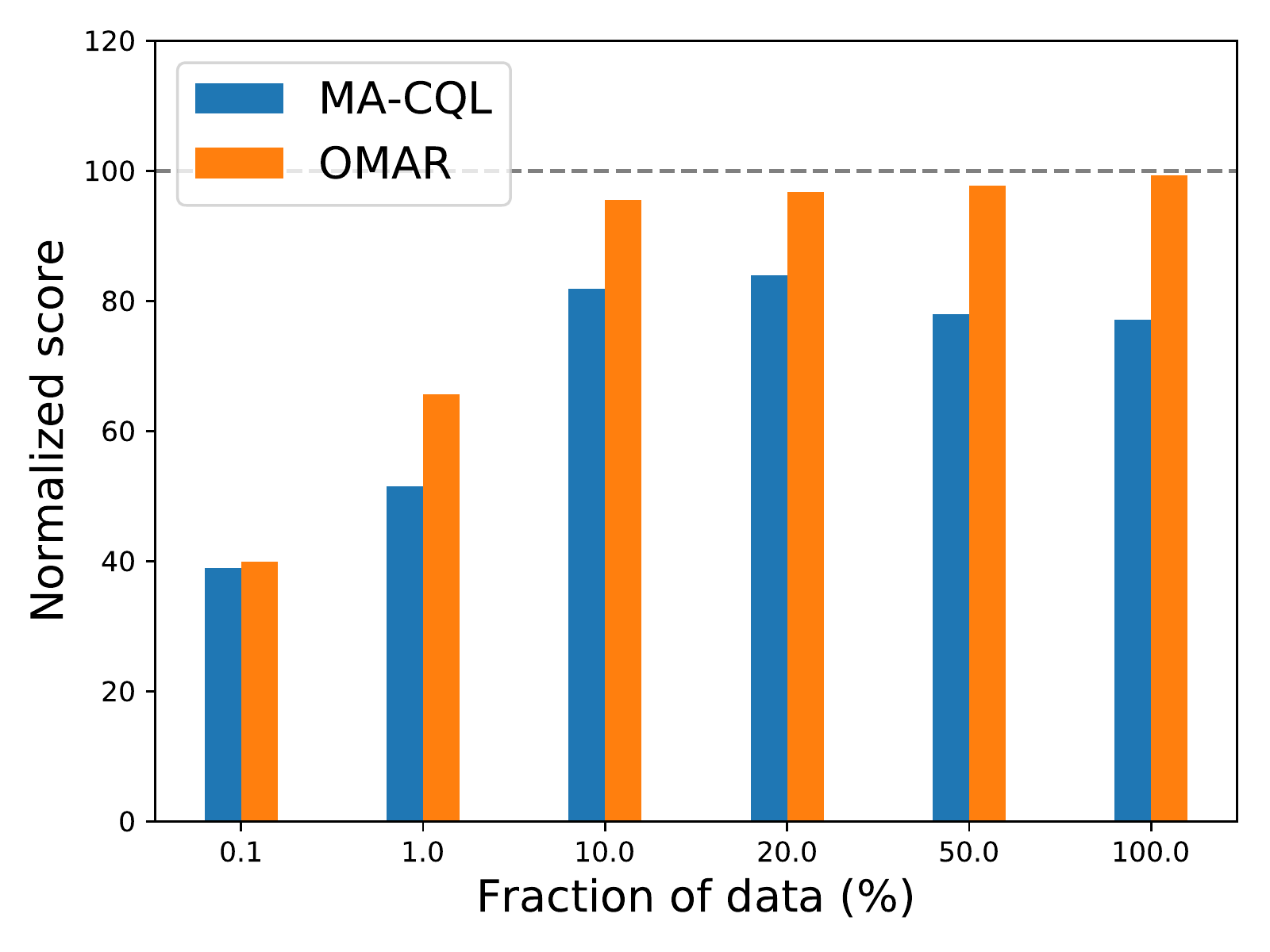}
\caption{Normalized score of OMAR and MA-CQL trained using a fraction of the entire replay dataset.}
\label{fig:size}
\end{figure}

Figure \ref{fig:size} shows that the performance of MA-CQL increases given more data points for $N\in\{1,10,20\}$. However, it does not further increase given an even larger amount of data, which performs much worse than the fully-trained online agents and fails to recover their performance. 
On the contrary, OMAR always outperforms MA-CQL by a large margin when $N>1\%$, whose performance is much closer to the fully-trained online agents given more data points. Therefore, the optimality issue still persists when dataset size becomes larger (e.g., it can take a very long time to escape from them if the objective contains very flat regions~\citep{ahmed2019understanding}). In addition, the zeroth-order optimizer part in OMAR can better guide the actor given a larger amount of data points with a more accurate value function. 

\subsection{Applicability on Centralized Training with Decentralized Execution}\label{sec:matd3}
\subsubsection{Results based on MATD3}
In this section, we demonstrate the versatility of the method and show that it can also be applied and beneficial to methods based on centralized critics under the CTDE paradigm.
Specifically, all baseline methods are built upon the MATD3 algorithm~\citep{ackermann2019reducing} using centralized critics as detailed in Section \ref{sec:maac}. Note that performance comparison and discussion of a centralized value function and a decentralized one is in Appendix \ref{app:ctde}.
Table \ref{tab:matd3_based} summarizes the averaged normalized score of different algorithms in each kind of dataset. As shown, OMAR (centralized) also significantly outperforms MA-ICQ (centralized) and MA-CQL (centralized), and matches the performance of MA-TD3+BC (centralized) in the expert dataset while outperforming it in other datasets. 

\begin{table}[!h]
\caption{The average normalized score of different methods based on MATD3 with centralized critics under the CTDE paradigm. 
}
\label{tab:matd3_based}
\centering
\begin{tabular}{ccccc}
\toprule
& Random & Medium-reply & Medium & Expert \\
\midrule
MA-ICQ & $5.2\pm5.5$ & $10.1\pm4.6$ & $27.4\pm5.3$ & $96.7\pm4.1$ \\
MA-TD3+BC & $7.9\pm2.2$ & $9.3\pm9.1$ & $29.4\pm3.7$ & $\textbf{108.1}\pm3.3$\\
MA-CQL & $12.8\pm4.9$ & $11.2\pm6.6$ & $26.3\pm13.3$ & $69.5\pm15.7$\\
OMAR & $\textbf{21.6}\pm4.6$ & $\textbf{19.1}\pm9.2$ & $\textbf{33.7}\pm14.5$ & $\textbf{105.9}\pm3.6$ \\
\bottomrule
\end{tabular}
\end{table}

\subsubsection{Discussion about the centralized and decentralized critics in offline multi-agent RL}\label{app:ctde}
We attribute the lower performance in Table \ref{tab:matd3_based} (based on a centralized value function) compared to Table \ref{tab:mpe} (based on a decentralized value function) due to the base algorithm, where Table \ref{tab:itd3_matd3} shows the performance comparison of offline independent TD3 and offline multi-agent TD3 in different types of dataset in cooperative navigation. 
As shown, utilizing centralized critics underperforms decentralized critics in the offline setting. 
There has also been recent research~\citep{de2020independent,lyu2021contrasting} showing the benefits of decentralized value functions compared to a centralized one, which leads to more robust performance. 
We attribute the performance loss of CTDE in the offline setting due to a more complex and higher-dimensional value function conditioning on all agent's actions and the global state that is harder to learn well without exploration.

\begin{table}[!h]
\caption{Averaged normalized score of ITD3 and MATD3 in cooperative navigation.}
\centering
\begin{tabular}{ccccc}
\toprule
& Random & Medium-replay & Medium & Expert \\
\midrule
ITD3 & $\textbf{18.7}\pm8.0$ & $\textbf{19.9}\pm4.7$ & $\textbf{18.6}\pm4.4$ & $\textbf{75.5}\pm7.9$ \\
MATD3 & $16.1\pm5.6$ & $12.7\pm6.1$ & $12.1\pm14.2$ & $1.6\pm2.7$ \\
\bottomrule
\end{tabular}
\label{tab:itd3_matd3}
\end{table}

\bibliography{main}
\bibliographystyle{plainnat}

\end{document}